\pdfoutput=1
\PassOptionsToPackage{table,dvipsnames}{xcolor}
\documentclass[10pt,twocolumn,letterpaper]{article}

\usepackage[pagenumbers]{cvpr}      %

\usepackage{multirow}
\usepackage{colortbl}
\usepackage{tabularx}
\usepackage{microtype}
\usepackage{pifont}
\usepackage{tikz}
\usepackage{subcaption}
\usetikzlibrary{arrows.meta,positioning,calc,fit,backgrounds,shapes.misc}

\usepackage[pagebackref,breaklinks,colorlinks]{hyperref}

\definecolor{accentblue}{HTML}{5D8DFD}
\definecolor{navy}{HTML}{3D4F7C}
\definecolor{mint}{HTML}{AFCFB1}
\definecolor{muted}{HTML}{777777}
\definecolor{panelbg}{HTML}{F6F7FB}
\definecolor{grid}{HTML}{DCDFE6}
\definecolor{plum}{HTML}{B7799E}
\definecolor{pink}{HTML}{DBBACB}
\definecolor{cream}{HTML}{EDE5E3}
\definecolor{steel}{HTML}{6E87B0}
\definecolor{bestred}{HTML}{D62728}
\definecolor{secondblue}{HTML}{1F5FBF}
\definecolor{slate}{HTML}{B2C1D6}
\colorlet{TableHeaderBg}{panelbg}
\colorlet{TableSubheadBg}{grid!42!white}
\colorlet{TableFocusBg}{plum!10!white}
\colorlet{GroupTxt}{navy}
\colorlet{GroupBg}{accentblue!10!white}
\colorlet{Besttxt}{bestred}
\colorlet{Secondtxt}{secondblue}
\hypersetup{linkcolor=navy,citecolor=navy,urlcolor=navy}

\newcommand{\TableHead}{\rowcolor{TableHeaderBg}}
\newcommand{\focusrow}{\rowcolor{TableFocusBg}}
\newcommand{\TableSetup}{\small\normalfont\setlength{\tabcolsep}{2.2pt}\renewcommand{\arraystretch}{1.10}}
\newcommand{\PrimaryGroup}[2]{\rowcolor{GroupBg}\multicolumn{#1}{l}{\textcolor{GroupTxt}{\textbf{\textit{#2}}}}}
\newcommand{\Best}[1]{\textcolor{Besttxt}{\textbf{#1}}}
\newcommand{\Second}[1]{\textcolor{Secondtxt}{\textbf{#1}}}
\newcommand{\HeadOne}[1]{\raisebox{0.26\baselineskip}[0pt][0pt]{#1}}
\DeclareRobustCommand{\ckY}{\ding{51}}
\DeclareRobustCommand{\ckN}{\textcolor{muted}{\ding{55}}}
\DeclareRobustCommand{\ckH}{\tikz[baseline=-0.6ex]{\draw[line width=0.4pt] (0,0) circle (0.55ex); \fill (0,-0.55ex) arc (-90:90:0.55ex) -- cycle;}}

\newcommand{\MarkLegend}{\ckY{}~provided\quad \ckH{}~partial\quad \ckN{}~absent}

\newcolumntype{Y}{>{\centering\arraybackslash}X}
\newcolumntype{L}{>{\raggedright\arraybackslash}X}
\newcolumntype{P}[1]{>{\raggedright\arraybackslash}p{#1}}
\newcolumntype{C}[1]{>{\centering\arraybackslash}p{#1}}
\newcommand{\method}{CaT-Movie\xspace}
\newcommand{\benchmark}{CaT-MBench\xspace}

\newcommand{\crossattnname}{shot-wise cross-attention\xspace}
\newcommand{\selfattnname}{reference-routed sparse self-attention\xspace}

\usepackage[most]{tcolorbox}
\usepackage[OT1,T1]{fontenc}
\usepackage{listings}
\usepackage{needspace}
\usepackage{placeins}
\newcommand{\ttdash}{{\fontencoding{T1}\selectfont\textendash}}

\newcommand{\RoleTag}[1]{\par\smallskip\noindent\textcolor{navy}{\textbf{\texttt{<#1>:}}}\enspace}

\newenvironment{PromptPlate}{%
  \par\addvspace{\baselineskip}\Needspace{8\baselineskip}\begingroup
  \captionsetup{type=table}%
}{\par\endgroup}
\tcbset{prompt/.style={
  enhanced,breakable,
  colback=white,colframe=grid,
  colbacktitle=TableHeaderBg,coltitle=navy,
  fonttitle=\small\bfseries,fontupper=\small\normalfont,
  boxrule=0.4pt,arc=0pt,
  left=8pt,right=8pt,top=6pt,bottom=6pt,
  before skip=6pt,after skip=6pt,
  pad at break*=4pt,
  before upper={\setlength{\parskip}{4pt}\setlength{\parindent}{0pt}\raggedright}
}}
\lstdefinestyle{PromptJSON}{
  basicstyle=\fontencoding{T1}\small\ttfamily,
  columns=fullflexible,keepspaces=true,
  breaklines=true,breakatwhitespace=false,breakindent=1.5em,
  showstringspaces=false,upquote=true,
  aboveskip=4pt,belowskip=4pt,
  xleftmargin=0pt,xrightmargin=0pt
}

\title{CaT-Movie: Balancing Global Planning and Local Rendering\\
for Long Multi-Shot Video Generation}

\author{
Jinpeng Lu\thanks{Equal Contribution.}\hspace{0.5em}
Haoyuan Shi\footnotemark[1]\hspace{0.5em} 
Yuanshen Guan\hspace{0.5em}
Jiyuan Xia\hspace{0.5em}
Bin Wu\\ 
Yinda Chen\hspace{0.5em} 
Zhiwei Xiong\thanks{Corresponding Author.}\\
\normalsize University of Science and Technology of China}
\hypersetup{
  pdftitle={CaT-Movie: Balancing Global Planning and Local Rendering for Long Multi-Shot Video Generation},
  pdfauthor={Jinpeng Lu, Haoyuan Shi, Yuanshen Guan, Jiyuan Xia, Bin Wu, Yinda Chen, Zhiwei Xiong}
}

\begin{document}
\maketitle

\begin{abstract}
Current multi-shot video generators render only limited durations per pass, requiring long movies to be stitched from local segments across which narrative progression and visual identity drift. To synthesize long-horizon movies with \textbf{C}oherence \textbf{a}cross \textbf{T}ime, we present \textbf{\method}, a multi-shot generation system consisting of a global story planner and a local window renderer. The \textbf{planner} expands user requests via a large language model to preserve, infer, and complete story details, while deterministic code maintains the resulting storyboard across shots to compile self-contained shot prompts and index reference frames from generated history.
To execute this guidance, the \textbf{renderer} applies \selfattnname and \crossattnname to route identities from the indexed frames while isolating per-shot textual conditions. 
Given the scarcity of high-quality open-source multi-shot data, we construct a dataset of over 350K film shots annotated with semantic state transitions, entity grounding, and multi-granularity requests. 
We further build \textbf{\benchmark}, a benchmark on held-out films with the same annotations. It gives every baseline its native format and counts cross-shot consistency only when a cut opens a new camera setup, preventing repeated framing from passing as consistency.
On \benchmark, \method attains the highest average score and the most consistent characters among all methods. The project page is available at \mbox{\urlstyle{same}\url{https://jinplu.github.io/CaT-Movie/}}.
\end{abstract}

\section{Introduction}
\label{sec:intro}
\begin{figure}[t]
  \centering
  \includegraphics[width=\columnwidth]{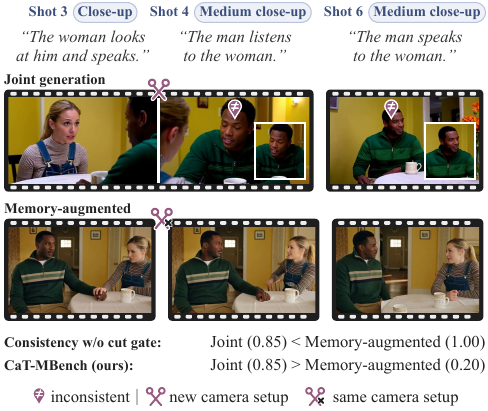}
  \caption{\textbf{Cross-window drift, copied framing, and cut-gated evaluation.} Each film strip is one generation pass. Joint generation switches camera setup at the cut from shot 3 to 4 but changes the man in the next pass, from shot 4 to shot 6, whereas memory-augmented generation keeps the man yet repeats the framing of shot 3 across planned cuts. Scores average character and scene consistency over all 11 shots, and the cut gate of \benchmark fails consistency checks across planned cuts that keep the camera setup.}
  \label{fig:teaser}
\end{figure}

Video diffusion transformers have enabled high-fidelity single-shot video synthesis from text~\cite{wan2025wan,wan2025wan22,gao2025seedance,seedance2026seedance2,klingteam2025klingomni}, and joint multi-shot generators now denoise several shots in one pass~\cite{guo2025long,qi2025mask2dit,meng2026holocine,wang2026multishotmaster,mao2026multigrid}. Long multi-shot videos such as movies, however, unfold their narrative states, visual identities, and scenes far beyond any generation window, as events leave lasting consequences and characters return with changed attributes. Because each pass renders only a limited duration, such videos are stitched from local segments in which every new shot rebuilds its scene, and narrative progression and visual identity drift across them~\cite{runway2025longer,google2025flow,lu2026current}.

Research systems automate parts of keeping \textbf{C}oherence \textbf{a}cross \textbf{T}ime in long videos, yet face a trade-off between cross-window consistency and camera transitions, which \cref{fig:teaser} illustrates. Joint generators coordinate shots and cuts within a window~\cite{meng2026holocine,wang2026multishotmaster,guo2025long} but lose consistency beyond it. Memory-augmented generators carry earlier frames forward, recalled by visual diversity, learned relevance, or named subjects~\cite{zhang2025storymem,an2025onestory,luo2025filmweaver,wang2026subjectanchor,vandersanden2026emvid}. A recalled frame, however, bundles a character's identity with the camera setup and story state of its shot, so recall not indexed by story state copies prior framing across planned cuts or returns stale appearances~\cite{yin2026evostate}. Agentic workflows plan the story before rendering it shot by shot~\cite{lin2023videodirectorgpt,wu2025movieagent,huang2026vimax,wang2026storyengine,wei2026complementary,xu2026reel}, yet drive frozen generators not trained on the plan format~\cite{wang2026storyengine,zuo2026filmworld}.

Existing benchmarks mostly obscure these failures by decoupling identity from camera cuts and story state~\cite{shi2026msvbench,chen2026cinedance,he2026entitybench,zeng2026cutcraft}. Without testing whether a planned cut opens a new place, shot size, angle, or cast, consistency checks reward dropped cuts or copied framing as in \cref{fig:teaser}, echoing \mbox{WorldScore}'s camera-control observation~\cite{duan2025worldscore}, while omitting explicit per-shot state continuity. Generation and evaluation thus miss the same record, a structured storyboard that separates who each entity is from the state and camera setup each shot requires, which a renderer can follow and a benchmark can check.

To this end, we present \textbf{\method}, which balances global story planning with local window rendering through a structured storyboard. Our \textbf{planner} uses a public language model to expand a request into registered entities and per-shot states and actions, preserving stated facts, inferring event consequences, and completing open details. Deterministic code maintains attire and held items across shots and compiles a self-contained prompt per shot. For each window, code queries the storyboard to recall generated anchor views of recurring characters and objects, preferring views after their latest planned attire or state change. Within each window, our \textbf{renderer} applies \crossattnname to isolate per-shot prompts and \selfattnname, so each shot reads only itself, the first latent frames of co-generated shots~\cite{meng2026holocine}, and recalled frames sharing its entities. Trained on storyboard-compiled prompts and same-scene shots often captured from another camera setup, it learns to borrow identity without copying framing.

Training and evaluating this system require films annotated in the same storyboard schema. Our shot-first pipeline processes over 350K film shots~\cite{chen2026cinedance,rawal2024cinepile,song2023moviechat}. Each scene block pairs multi-granularity requests with an entity registry, per-shot states, within-shot transitions, and entity grounding from boxes, masks, and depth~\cite{carion2025sam3,wang2026vggtomega}. From held-out films in the same schema, we build \textbf{\benchmark} to test what the storyboard plans. Its questions ask whether each planned cut opens a new camera setup, whether returning entities and scenes keep their identity, and whether state changes are established and persist. Consistency across a planned cut counts only when the video opens the new setup, denying copied framing any credit. On \benchmark, \method executes 0.974 of planned cuts and attains the highest average score and the most consistent characters, at 0.651.

In summary, our contributions are fourfold.
\begin{itemize}
    \item \textbf{Global story planner.} A training-free planner pairing language-model expansion with code maintenance to compile shot prompts and index anchor views.
    \item \textbf{Local window renderer.} A multi-shot renderer with \crossattnname, \selfattnname, and cross-setup scene training that preserves identity across windows without copying framing.
    \item \textbf{Film storyboard dataset.} Over 350K film shots annotated in a shot-first pipeline with entity registries, state transitions, entity grounding, and multi-granularity requests. 
    \item \textbf{Cut-gated story benchmark \benchmark.} Held-out film stories with native-format baseline adapters and gated checks of consistency and state continuity.
\end{itemize}

\section{Related Work}
\label{sec:related}
\begin{figure*}[t]
  \centering
  \includegraphics[width=\textwidth]{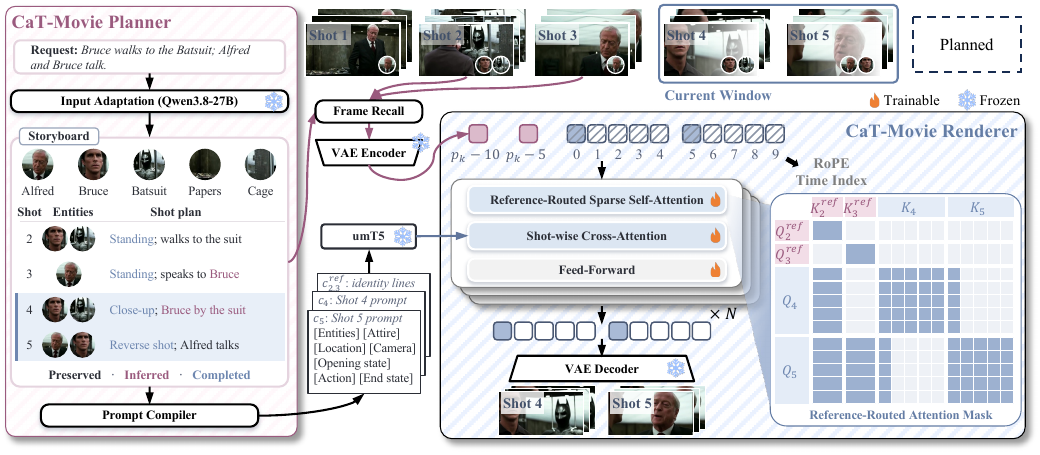}  \caption{\textbf{Overview of \method.} A planner writes one storyboard, from which a prompt compiler maintains cross-shot state and derives per-shot text $c_k$, the frames each shot recalls, and an identity line $c^{\mathrm{ref}}$ for each recalled frame. The renderer denoises each window jointly with \selfattnname and \crossattnname, placing recalled frames before the reading shot's first latent frame.}
  \label{fig:overview}
\end{figure*}

\paragraph{Multi-shot video generation and visual memory.}
Joint generators denoise several shots in one window through masked attention, boundary or planning tokens, shot-aware rotary phases, or spatial grids~\cite{guo2025long,qi2025mask2dit,wu2026cinetrans,kara2025shotadapter,guo2026shotplan,wang2025echoshot,wang2026multishotmaster,mao2026multigrid}. Wide cross-shot access leaks content and weakens transitions~\cite{wu2025shotdirector}, motivating sparse designs that exchange only through per-shot summary tokens such as first latent frames~\cite{meng2026holocine}. To reach beyond a single window, keyframe pipelines animate consistent images~\cite{zhou2024storydiffusion,huang2024iclora,xiao2025captaincinema}, memory methods recall earlier generated frames~\cite{zhang2025storymem,an2025onestory,wei2026memento,wu2026cinemem,luo2025filmweaver,xu2026reel,wang2026weaveforcing,vandersanden2026emvid,wang2026subjectanchor}, and reference models inject given images~\cite{xue2025standin} or bind them to target shots~\cite{wang2026multishotmaster,huang2026cineweaver,huang2026poco}. Yet similarity- or recency-based recall drifts over multiple windows or copies prior framing~\cite{luo2025filmweaver,vandersanden2026emvid}. Our renderer routes storyboard-indexed anchor views by shared entities with explicit identity lines and trains on same-scene shots from different camera setups to preserve identity without copying framing.

\paragraph{Story planning and prompt enhancement.}
Because a recalled frame may show an outdated appearance~\cite{yin2026evostate,chen2026morememory,lu2026current}, both recalled views and shot prompts must track story progression. Agentic pipelines plan multi-shot stories~\cite{lin2023videodirectorgpt,wu2025movieagent,huang2026vimax}, tracking planned world states~\cite{wang2026storyengine}, keying assets to costume states~\cite{zuo2026filmworld}, writing back states parsed from generated video~\cite{yin2026evostate,miao2026do}, or, in concurrent work, selecting generated references~\cite{wei2026complementary,xu2026reel}, yet their plans drive frozen generators not trained on the plan format. Prompt enhancers write structured captions matched to generator training but only for a single pass~\cite{zhu2026wanpe,chen2026scaling}. Our planner bridges both: a language model expands requests into our renderer's storyboard schema, and code maintains cross-shot state to compile per-shot prompts and index post-change anchor views.

\paragraph{Multi-shot datasets and story evaluation.}
Multi-shot datasets caption clips shot by shot~\cite{meng2026holocine,wang2026multishotmaster,chen2026cinedance,zhang2026muss}, and concurrent datasets add entity states~\cite{zhou2026dramadirector,zhao2026ssrs} or 3D person geometry~\cite{zou2026miseenvideo}, yet none combines a cross-shot entity registry, typed per-shot states, within-shot transitions, and entity grounding, as App.~D details. Benchmarks score video quality, events, story images, editing, audio, or consistency~\cite{huang2024vbench,zheng2025vbench,wang2025your,feng2025tc,zhuang2025vistorybench,zeng2026cutcraft,liu2026castbench,shi2026msvbench,he2026entitybench}, and those tying consistency to cuts do so only at adjacent boundaries~\cite{wu2026cinemem} or per video~\cite{yang2026adspark}. Our dataset provides all four annotations with multi-granularity requests, and \benchmark gates consistency between any two shots on opening a new camera setup whenever the storyboard plans one.

\section{Method}
\label{sec:method}
\subsection{Problem Formulation and Overview}
\label{sec:setting}
Given a story request, written as a one-sentence synopsis, a scene paragraph, or a numbered shot list, and a target duration, we generate a multi-shot video. A diffusion pass renders a limited duration, 30\,s even for the commercial Seedance 2.5~\cite{seed2026seedance25}. A long story is therefore generated in windows of consecutive shots, which we cap at 8\,s in training under our compute budget and at 7\,s on the test stories. As \cref{fig:overview} shows, a planner writes one storyboard $S$ before video generation, code derives from $S$ the self-contained prompt $c_k$ of every shot $k$ and the frames $R_k$ it recalls from earlier windows, and a window renderer denoises each window jointly in story order.

\subsection{Storyboard Planner}
\label{sec:planner}
\label{sec:storyboard}

\begin{figure*}[t]
  \centering
  \includegraphics[width=\textwidth]{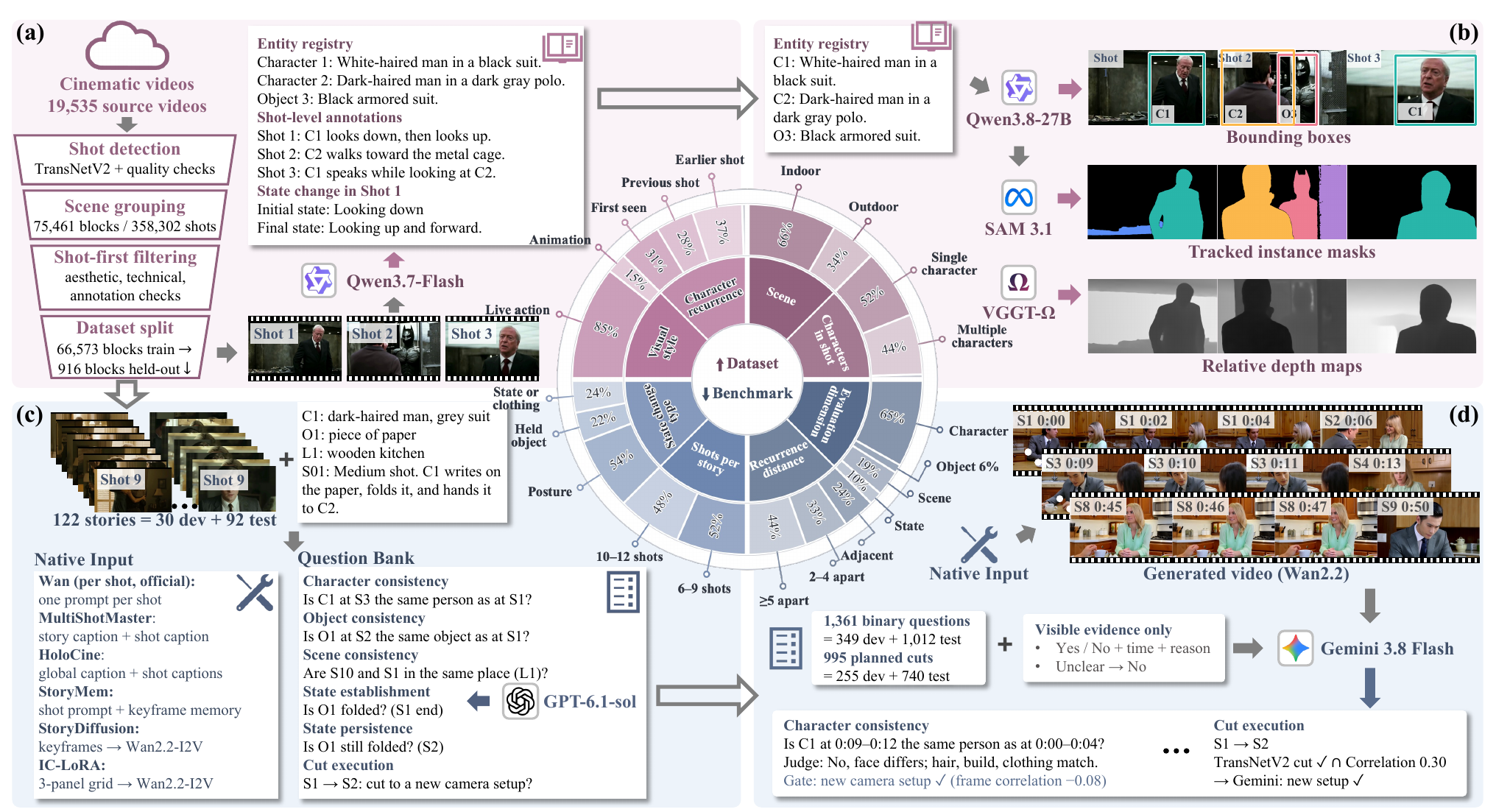}
  \caption{\textbf{Overview of the dataset and benchmark.} One film pool and one schema feed the training data, (a) curation and annotation and (b) entity grounding, and \benchmark, (c) stories and questions and (d) cut-gated judging. The center chart gives the composition of both.}
  \label{fig:infra}
\end{figure*}

Single-pass prompt enhancers~\cite{zhu2026wanpe} do not coordinate prompts across windows. Our planner instead uses Qwen3.8-27B~\cite{qwen2026qwen38} without fine-tuning to write one structured storyboard that deterministic code maintains and queries across shots.

\paragraph{Structured storyboard.}
A storyboard $S$ pairs an entity registry with a plan for every shot. The registry assigns each recurring character, object, and location a fixed identifier and identity description, plus a default attire per character. The plan of shot $k$ records the location, the camera's shot size, angle, and movement, an optional opening composition, the action, and a suggested shot duration. It lists the present entities $\mathcal{E}_k$ with their opening visible state and each character's posture, held items, and attire, and writes within-shot changes with initial and visible final values, such as held items ending \emph{empty-handed}.

\paragraph{Input adaptation.}
Requests, evaluated in \cref{sec:exp_planner}, range from shot lists fixing per-shot details to synopses leaving most open. Since generation fidelity scales with structured visual information~\cite{chen2026scaling}, one unified prompt (App.~B) instructs the model across all granularities to preserve stated facts, infer implied presence and event outcomes, and complete open details such as postures. Avoiding invented staging, it writes a composition only when the request implies an opening layout, keeps stated action positions, and leaves the remaining layout to the renderer. Constrained decoding guarantees valid JSON, and a plan that misses a shot list's shot count or has too few shots for the duration is sent back once.

\paragraph{Maintenance and compilation.}
Code, the prompt compiler in \cref{fig:overview}, copies back the entity descriptions and camera settings a shot list states and propagates deterministic state across $S$: it carries each character's attire from its latest recorded change into later shots and fills unstated held items from the end of that character's previous shot. Recording when changes occur stays with the model, along with posture and visible state, which mix lasting marks with fleeting expressions. Code scales shot durations to match the target duration and packs consecutive shots greedily into windows under the cap, a longer shot filling a window alone. The plan of each shot $k$ compiles into a self-contained prompt $c_k$ restating present entity descriptions, current character attire, and change outcomes. Because our renderer trains on prompts from the same compiler and $S$ also indexes the recalled frames $R_k$ of \cref{sec:renderer}, every shot's text and visual references stem from one state record.

\subsection{Window Renderer}
\label{sec:renderer}
\label{sec:analysis}
\label{sec:training}

Without visual history, per-shot generation loses character identity across shots, as \cref{sec:main} shows, yet recalled frames bundle persistent identity with potentially outdated states and framing that should not be copied~\cite{chen2026morememory}. Our renderer, Wan2.2-TI2V-5B~\cite{wan2025wan22} fine-tuned without new parameters, borrows identity from recalled frames while taking state and framing from each shot's prompt.

\paragraph{Recalling generated frames.}
Unlike similarity-based retrieval~\cite{zhang2025storymem,an2025onestory}, recall is a deterministic query on $S$. Because recalling an entity's latest appearance chains errors across windows, each entity keeps its clearest view as an \emph{anchor view} while its planned appearance holds (App.~A). Consider a target shot $k$ in a window starting at shot $k_0$ and an earlier shot $j<k_0$ containing a present character or object $e\in\mathcal{E}_k$. Let $u_j\in\{0,1\}$ equal $1$ if shot $j$ precedes the latest attire or visible-state change recorded for $e$ before shot $k$, $n_j$ count other registered characters in shot $j$, and $d_j$ rank shot sizes by character legibility, close-up, medium close-up, and medium shots first, with $d_j=0$ for objects. Entity $e$ selects
\begin{equation}
  \rho_k(e)=\operatorname*{arg\,min}_{j<k_0,\ e\in\mathcal{E}_j}\big(u_j,\ n_j,\ d_j,\ j\big),
  \label{eq:recall}
\end{equation}
compared lexicographically. Each target shot also selects the shot before its window at its location with the fewest characters, then the widest shot size, then the earliest index. Keeping up to ten unique source shots per window, we extract each shot's last frame if it records any attire or visible-state change and its middle frame otherwise. Within the window, $R_k$ holds the recalled frames whose source shot shares a present entity with shot $k$, preventing shots from borrowing framing from frames that show only other characters. Frames sharing no entity with the window, such as empty location views, enter every $R_k$.

\paragraph{Window attention.}
Each shot is VAE-encoded separately so no latent frame spans a cut. Let $Z_k$ denote the latent tokens of shot $k$ in window $\mathcal{W}$, $Z_k^{\mathrm{f}}$ its first latent frame, and $Z^{\mathrm{ref}}_r$ the clean latents of recalled frame $r$. Under \crossattnname, $Z_k$ attends only to $c_k$~\cite{meng2026holocine}. Under \selfattnname, it computes
\begin{equation}
  \mathrm{Attn}\big(Z_k,\,[\,Z_k,\ \{Z_j^{\mathrm{f}}\}_{j\in\mathcal{W},\,j\neq k},\ \{Z^{\mathrm{ref}}_r\}_{r\in R_k}\,]\big),
  \label{eq:window_attention}
\end{equation}
attending to its own shot, the first latent frames of co-generated shots as compact summaries~\cite{meng2026holocine}, and its routed references $R_k$~\cite{wang2026multishotmaster,huang2026poco}. Recalled frames keep the height--width RoPE grid of the shots and, for each reading shot $k$, take temporal positions five apart ending at $p_k-5$, where $p_k$ is the position of $Z_k^{\mathrm{f}}$, rather than acting as adjacent preceding frames~\cite{zhang2025storymem,xue2025standin}. Each attends only to itself and cross-attends to an identity line $c^{\mathrm{ref}}_r$ marking it as another shot and naming, in the identifiers of $c_k$, the cast and setting of its source shot, from frame labels in training and from $S$ at inference.

\paragraph{Training and inference.}
We fine-tune all transformer blocks with the rectified flow objective~\cite{lipman2023flow,liu2023flow} on each window $\mathcal{W}$,
\begin{equation}
  \mathcal{L}=\mathbb{E}_{t,\epsilon}\sum_{k\in\mathcal{W}}\big\|\hat{v}_k-(\epsilon_k-Z_k)\big\|_2^2,
  \label{eq:objective}
\end{equation}
where $\epsilon_k$ is Gaussian noise and $\hat{v}_k$ predicts the velocity of shot $k$ from the noised latents $Z_k^{t}=(1-t)Z_k+t\epsilon_k$ at timestep $t$ during joint denoising, with no loss on references. Training windows pair whole shots from one film scene with one frame each from one to ten other shots of that scene, entity-sharing shots first. These frames often show the same cast under another camera setup, so copying their framing raises $\mathcal{L}$, matching the cut gate of \cref{sec:benchmark}, and random zoom-crops on about half of them vary framing further. Training references receive light noise to narrow the gap to generated frames and drop together with probability 0.2 for history-free openings, while inference uses clean ones from \cref{eq:recall}.

\section{Dataset \& Benchmark}
\label{sec:data_bench}

\subsection{Film Storyboard Dataset}
\label{sec:data}

Training the renderer and testing the system need films annotated in the storyboard schema, which no existing multi-shot dataset provides in full, as App.~D compares. We split 19,535 videos from CineDance~\cite{chen2026cinedance}, CinePile~\cite{rawal2024cinepile}, and MovieChat~\cite{song2023moviechat} into shots and group consecutive shots of one scene into blocks of up to twelve, totaling 75,461 blocks with 358,302 shots and 332.8 hours. Qwen3.7-Flash annotates each block in one pass with the storyboard of \cref{sec:storyboard}, which keeps identifiers consistent across its shots, together with a scene paragraph and a synopsis. Those two tiers and the compiled shot list give planner requests at three granularities. Aesthetic~\cite{schuhmann2022laion}, technical~\cite{wu2023dover}, and annotation checks drop single shots rather than every window containing one, which retains 28.1\% more usable shots, and the 67,489 blocks that keep a usable shot are split by film into 66,573 for training and 916 held out. Entity boxes, masks, and depth from Qwen3.8-27B, SAM~3.1~\cite{carion2025sam3}, and VGGT-$\Omega$~\cite{wang2026vggtomega} give the spatial relations in training text, as \cref{fig:infra}a,b summarizes.

\subsection{\benchmark}
\label{sec:benchmark}

Existing multi-shot evaluations score cross-shot consistency as feature similarity between shots, ViCLIP~\cite{wang2024internvid} for HoloCine and StoryMem~\cite{meng2026holocine,zhang2025storymem} and DINOv2~\cite{oquab2024dinov2} on subject crops for MultiShotMaster and OneStory~\cite{wang2026multishotmaster,an2025onestory}, whether or not a planned cut happened, so repeating the previous framing raises the score. \benchmark instead makes three choices: a judge asks whether each registered entity is the same individual when it returns; this check counts only across cuts that open the planned new camera setup; and a requested change must be made in its shot and persist on return, which no existing benchmark pairs with per-shot states (App.~E).

\paragraph{Construction.}
We build 122 stories from held-out films that no training block uses. Each has six to twelve shots and at most 60\,s, a within-shot change, a character returning across non-adjacent shots, and a dynamic action shot, and 30 stories are for development and 92 for testing. The baselines of \cref{sec:main} render the reference storyboard annotated from the film, given in each one's native prompt format with every storyboard field, as App.~F shows for one shot. As \cref{fig:infra}c illustrates, GPT-6.1-sol lists the entities, actions, and changes of each shot-list request, and a program turns the list into binary questions that name entities by registry identifier. The test set has 1,012 questions and 740 planned cuts, 674 of which change the place, shot size, angle, or cast.

\paragraph{Metrics.}
Execution, consistency, and state scores are pass rates of binary checks, averaged within a story and then over stories. \emph{Transition control} averages \emph{Cut F1} over all 740 planned cuts, matching detected cuts within 0.125\,s, and \emph{Execution}, the executed share of the 674. HoloCine misses this tolerance at 175 cuts by rounding them to its latent grid. \emph{Inter-shot consistency} checks that a character, an object, or the scene matches its first appearance when it returns. \emph{State continuity} checks that a requested change is made in its shot, \emph{Establish}, and still holds when the entity returns, \emph{Persist}. \emph{Text alignment} is the ViCLIP similarity of each shot to its request line. Gemini 3.8 Flash answers the consistency, state, and camera-setup checks, as in \cref{fig:infra}d. Each of the 674 cuts counts as executed only if TransNetV2 detects a cut within 0.5\,s and the judge finds a new camera setup, and the same test gates every consistency and Persist check between shots the storyboard makes different. A check behind a failed gate fails and stays in the average. App.~E gives the thresholds and prompts.

\begin{table*}[t]
  \centering
  \caption{\textbf{Quantitative comparison} on 92 test stories. \method renders its planner's storyboard from the shot-list request, which states less than the reference storyboard the baselines render. Avg.\ averages the four aspects, with \Best{best} and \Second{second-best} marked.}
  \label{tab:main}
  \TableSetup
  \begin{tabularx}{\textwidth}{ll!{\color{grid}\vrule}Yc!{\color{grid}\vrule}YYY!{\color{grid}\vrule}YY!{\color{grid}\vrule}c!{\color{grid}\vrule}Y}
    \toprule
    \TableHead
    & & \multicolumn{2}{c}{Transition Control$\uparrow$} & \multicolumn{3}{c}{Inter-shot Consistency$\uparrow$} & \multicolumn{2}{c}{State Continuity$\uparrow$} & Text$\uparrow$ & \\
    \TableHead
    \HeadOne{Method} & \HeadOne{Video base} & Cut F1 & Execution & Character & Object & Scene & Establish & Persist & Alignment & \HeadOne{Avg.$\uparrow$} \\
    \midrule
    \PrimaryGroup{11}{Per-shot generation} \\
    Wan2.1, per shot & Wan2.1-T2V-1.3B & \Second{0.998} & \Second{0.994} & 0.219 & 0.169 & 0.310 & 0.274 & 0.694 & 0.181 & 0.474 \\
    Wan2.2, per shot & Wan2.2-TI2V-5B & \Best{0.999} & \Best{0.996} & 0.227 & 0.275 & 0.310 & 0.387 & \Second{0.701} & 0.174 & 0.497 \\
    StoryMem & Wan2.2-I2V-A14B & 0.928 & 0.719 & \Second{0.631} & 0.453 & \Best{0.687} & 0.484 & 0.545 & 0.179 & 0.527 \\
    StoryDiffusion & Wan2.2-I2V-A14B & 0.905 & 0.825 & 0.300 & 0.265 & 0.245 & \Second{0.597} & 0.535 & 0.171 & 0.468 \\
    IC-LoRA & Wan2.2-I2V-A14B & 0.976 & 0.961 & 0.168 & 0.133 & 0.406 & 0.452 & 0.663 & 0.176 & 0.484 \\
    \PrimaryGroup{11}{Joint multi-shot generation} \\
    MultiShotMaster & Wan2.1-T2V-1.3B & 0.994 & 0.990 & 0.538 & 0.254 & 0.582 & 0.371 & 0.587 & 0.178 & 0.527 \\
    HoloCine & Wan2.2-T2V-A14B & 0.748 & 0.967 & 0.517 & \Best{0.512} & 0.586 & \Best{0.694} & 0.691 & \Best{0.195} & \Second{0.571} \\
    \focusrow \method & Wan2.2-TI2V-5B & 0.995 & 0.974 & \Best{0.651} & \Second{0.497} & \Second{0.630} & 0.468 & \Best{0.785} & \Second{0.186} & \Best{0.597} \\
    \bottomrule
  \end{tabularx}

\end{table*}

\paragraph{Validation against human annotation.}
Two annotators labeled question--video items from seven baselines, blind to generating method and judge answers, and a reviewer resolved all disagreements from video frames before viewing labels, leaving 557 items. Against reviewed ground truth, the judge achieves Cohen's $\kappa=0.70$, lowest on scenes where a reverse shot reveals an unseen side of the location, while the annotators' pooled blind labels reach $\kappa=0.85$ in App.~E. On these labels, ViCLIP has an AUC of 0.49--0.70.

\section{Experiments}
\label{sec:experiments}
\begin{figure}[t]
  \centering
  \includegraphics[width=\columnwidth]{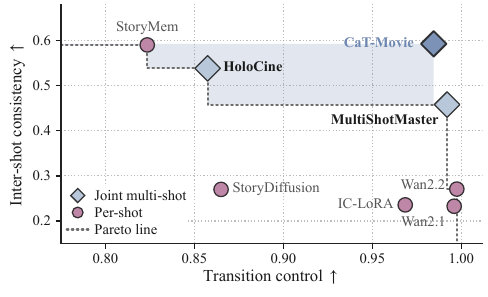}
  \caption{\textbf{Inter-shot consistency against transition control} on the 92 test stories. The dotted line marks the baseline Pareto frontier, and the blue region lies beyond every baseline.}
  \label{fig:tradeoff}
\end{figure}

\begin{figure}[t]
  \centering
  \includegraphics[width=\columnwidth]{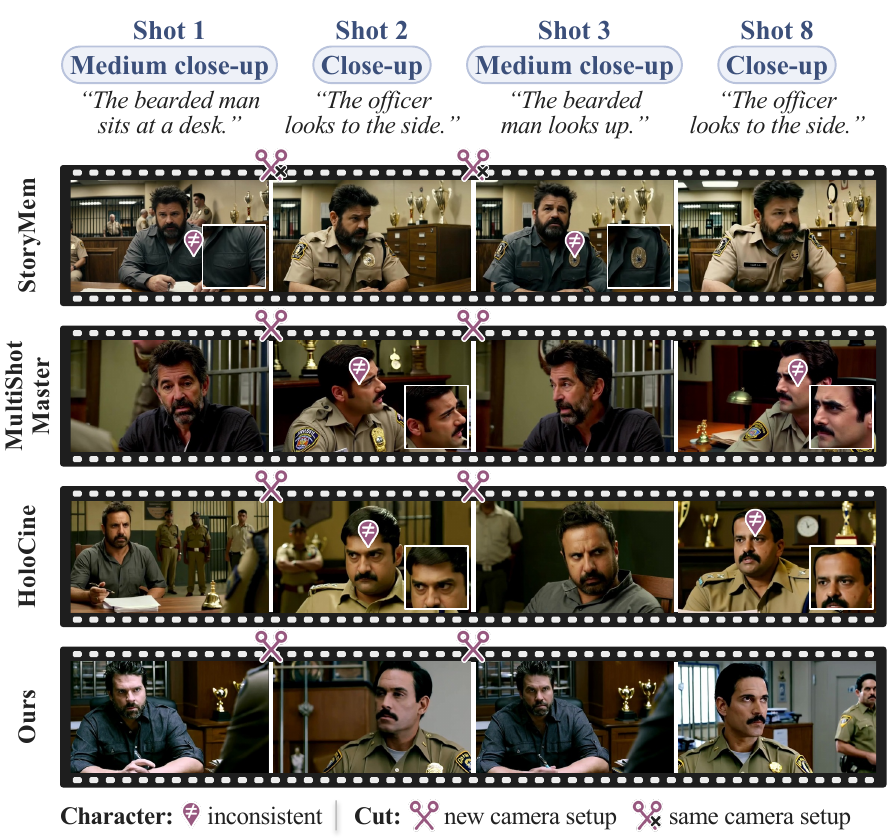}
  \caption{\textbf{Qualitative comparison on a test story of 11 shots and 28.4\,s}, which HoloCine generates in three windows. Each of four shots is shown at its middle frame. Icons mark characters that \benchmark judges inconsistent, enlarged at the bottom right of the shot, and crossed scissors mark planned cuts where the video keeps the camera setup.}
  \label{fig:qualitative}
\end{figure}

\label{sec:impl}

We compare against seven baselines: per-shot Wan2.1-T2V-1.3B~\cite{wan2025wan} and Wan2.2-TI2V-5B~\cite{wan2025wan22}, multi-shot MultiShotMaster~\cite{wang2026multishotmaster} and HoloCine~\cite{meng2026holocine}, and memory-based StoryMem~\cite{zhang2025storymem}. Two keyframe pipelines pair StoryDiffusion~\cite{zhou2024storydiffusion} on SDXL~\cite{podell2023sdxl} or In-Context LoRA~\cite{huang2024iclora} on FLUX.1-dev~\cite{bfl2024flux}, termed IC-LoRA, with Wan2.2-I2V-A14B~\cite{wan2025wan22i2v}.
\method renders the test stories with a 7\,s window cap, the cap whose window lengths are closest to those of training. The renderer is trained for 3,000 steps on 8 A100 GPUs (App.~A; code entry points in App.~H). Renderer ablations in \cref{tab:ablation_renderer} use the 8\,s training cap on the reference storyboards of the development stories. Differences are in points, that is scores times 100, and a difference is significant when its paired 95\% interval over the same stories excludes zero, from 10,000 resamples of source films with seeds averaged per story.

\subsection{Main Comparison}
\label{sec:main}
\label{sec:longrange}

\paragraph{Benchmark findings.} Every method generated all 92 test stories in \cref{tab:main}, and across the baselines transition control and inter-shot consistency trade off in \cref{fig:tradeoff}.

\textbf{Finding 1: Joint generation keeps identity only within a window.} HoloCine and MultiShotMaster execute nearly all planned cuts but keep characters at only 0.517 and 0.538. Their character checks fall from 58--61\% for returns within 10\,s to 41--49\% later in \cref{fig:window_distance}~(left), and the window boundary rather than elapsed time accounts for this: within one of their windows the checks pass 62--83\% at every gap, while across windows they pass significantly fewer, 28\% and 30\%, even 3--10\,s apart, like HoloCine's officer in \cref{fig:qualitative}.

\textbf{Finding 2: Memory-based consistency partly comes from copying camera setups.} StoryMem, the most consistent baseline, cuts on time (Cut F1 0.928) but repeats the framing at 114 of all 674 planned setup changes, as in \cref{fig:teaser,fig:qualitative}. The judge finds 124 of the 145 consistency checks behind its failed gates consistent (App.~E), so without the gate StoryMem would lead every method by a significant 15.3 points or more in consistency, a lead the gate removes. The gate fails 15.7\% of its consistency checks against 0.7\% for \method, and ViCLIP similarity rates StoryMem 8.4 points above the source films.

\textbf{Finding 3: Independent per-shot generation cuts cleanly but forgets.} Cutting between separately generated shots, the Wan baselines exceed 0.99 in Cut F1 and Execution but keep characters at only 0.22 and 0.23, and the two keyframe pipelines, which animate each keyframe separately, at 0.30 and 0.17.

\paragraph{Breaking the trade-off.} \method is the only method beyond the baseline Pareto line in \cref{fig:tradeoff}. It executes 0.974 of planned cuts, on par with joint generators, reaches the highest inter-shot consistency of 0.593, and keeps characters consistent at 0.651, significantly above every baseline but StoryMem, which it matches while executing significantly more cuts. It matches HoloCine within HoloCine's windows, leads it by 22.2 points across them, and passes 61--63\% of character checks at every return time in \cref{fig:window_distance}~(left). In \cref{fig:qualitative}, it keeps the officer across windows and opens both planned setups. Its average of 0.597 is the highest, 2.6 points above HoloCine and significantly above every other baseline. Under the 7\,s window cap, 617 of 675 entity returns in our storyboards cross a window boundary.

\paragraph{State inheritance.} \method also keeps the most requested changes when entities return, with Persist of 0.785, significantly above StoryMem by 24.0 points and MultiShotMaster by 19.8 points. This points to the storyboard text rather than recalled frames: training without references does not lower Persist in \cref{tab:ablation_renderer}, while with the same renderer our planner's storyboard raises it over the reference storyboard and the request (\cref{sec:exp_planner}). Establish instead follows the video base more than the paradigm, from 0.27 and 0.37 on Wan2.1-1.3B to 0.45--0.69 for the four methods on Wan2.2 A14B, and \method lifts its 5B base from 0.387 to 0.468.

\subsection{Ablations}
\label{sec:exp_ablation}
\begin{table}[t]
  \centering
  \caption{\textbf{Ablations} on 30 development stories over three seeds. \method renders the reference storyboard, the first group retrains from the base model, and the last replaces it with the shot-list request compiled directly or our planner's storyboard. Avg.\ also counts Cut F1, Establish, and text alignment, and $^\dagger$ marks a significant difference from \method.}
  \label{tab:ablation_renderer}
  \TableSetup
  \begin{tabularx}{\columnwidth}{l *{6}{Y}}
    \toprule
    \TableHead
    Setting & Exec. & Char. & Obj. & Scene & Persist & Avg. \\
    \midrule
    \focusrow \method & 0.958 & 0.581 & 0.585 & 0.566 & 0.615 & 0.571 \\
    \PrimaryGroup{7}{Trained without} \\
    \quad identity lines & 0.942 & 0.582 & 0.607 & 0.557 & 0.653 & 0.561 \\
    \quad routing & 0.923$^\dagger$ & 0.593 & 0.513 & 0.595 & 0.609 & 0.564 \\
    \quad references & 0.966 & 0.289$^\dagger$ & 0.342$^\dagger$ & 0.474 & 0.696 & 0.521$^\dagger$ \\
    \PrimaryGroup{7}{Reference selection} \\
    \quad recent shots & 0.928$^\dagger$ & 0.488$^\dagger$ & 0.615 & 0.471$^\dagger$ & 0.598 & 0.563 \\
    \quad CLIP retrieval & 0.953 & 0.544 & 0.538 & 0.503 & 0.534 & 0.537$^\dagger$ \\
    \quad + first views & 0.955 & 0.569 & 0.547 & 0.546 & 0.667 & 0.566 \\
    \PrimaryGroup{7}{Storyboard} \\
    \quad request only & 0.955 & 0.525 & 0.509 & 0.546 & 0.557 & 0.541 \\
    \quad our planner & 0.963 & 0.579 & 0.530 & 0.560 & 0.708 & 0.587 \\
    \bottomrule
  \end{tabularx}
\end{table}

\paragraph{Renderer.}
In \cref{tab:ablation_renderer}, training without identity lines leaves consistency unchanged but significantly raises copying from 3.6\% to 5.0\% of the shot pairs planned to change camera setup, counting pairs judged the same setup or not cut, so naming who each recalled frame shows helps the renderer take identity without framing. Training without routing, where every shot reads every recalled frame of its window, significantly loses 3.5 points of cut execution. The loss sits in cuts where routing would hide recalled frames showing only characters absent from the shot: their execution falls from 0.970 to 0.869, with 32 of the 34 failures copying a framing, while other cuts change by 1.4 points. Copying a recalled frame's framing is thus a shortcut, the kind of input dependence that behavior tracing exposes in vision-language-action models~\cite{shi2026vlatrace}. Training without references keeps cut execution but loses 29.2 and 24.4 points of character and object consistency.

\paragraph{Reference selection.}
The recent-shots and CLIP rows keep the renderer and the number of recalled frames. Taking frames from the latest generated shots instead of querying the storyboard significantly loses 3.0, 9.3, and 9.5 points of cut execution, character consistency, and scene consistency. Retrieving them by CLIP image--text similarity to the shot text, the retrieval baseline of Comp-RAP~\cite{wei2026complementary}, significantly lowers Avg.\ by 3.4 points. Adding each entity's earliest post-change view to the anchor views changes nothing significantly.

\begin{figure}[t]
  \centering
  \includegraphics[width=\columnwidth]{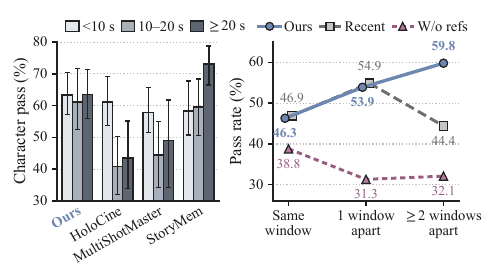}
  \caption{\textbf{(Left)~Character checks passed by time from first appearance to return}, pooled over the test stories, with 95\% intervals. \textbf{(Right)~Consistency by window distance} on development stories, where Recent is recent-shots recall and W/o refs trains without references.}
  \label{fig:window_distance}
\end{figure}

\paragraph{Beyond the window.}
Under greedy packing, 84\% of the consistency pairs on the development stories compare shots in different windows. Grouped by window distance in \cref{fig:window_distance}~(right), training without references loses 22.6 and 27.7 points one and two or more windows apart, and the recent-shots row loses 15.4 points only two or more windows apart. On character checks there, recent-shot selection still shows the character in 91\% of them but recalls a frame from the window of its first appearance in only 22\%, against 60\% for the storyboard, and it loses where it recalls a later regenerated appearance, the error chaining that anchor views avoid. Consistency across windows thus comes from storyboard-indexed recalled frames, and no window cap or packing in App.~A changes any column significantly.

\subsection{Storyboard Planner Evaluation}
\label{sec:exp_planner}
\label{sec:exp_paradigm}
\begin{figure}[t]
  \centering
  \includegraphics[width=\columnwidth]{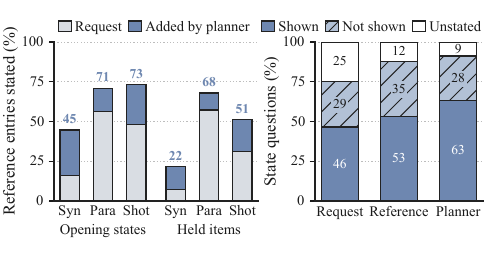}
  \caption{\textbf{Planner text and rendering.} \textbf{(Left)}~Opening states and held items of the reference storyboard stated by the request alone and with our planner, for synopsis, paragraph, and shot-list requests marked Syn, Para, and Shot. \textbf{(Right)}~State questions of the storyboard rows of \cref{tab:ablation_renderer} whose change the per-shot text states and the video shows, states but the video omits, or leaves unstated.}
  \label{fig:planner_text_main}
\end{figure}

Following the evaluation of prompt enhancers~\cite{zhu2026wanpe}, we check the planner's text on the synopsis, paragraph, and shot-list requests of the 30 development stories and of 40 unseen stories from films outside the benchmark, with two planner seeds and one prompt frozen on the development stories before the test stories were planned. A text-only judge, DeepSeek-V4.1-Flash, checks the compiled per-shot text against the events and outcomes each request requires, written before any plan is read, and against the reference storyboard, as App.~C details. We then render the 30 development stories with the same renderer from the request compiled directly, our planner's storyboard, and the reference storyboard.

\paragraph{Information in the text.}
Synopses state 16\% of the reference storyboard's opening states and 7\% of its held items, and shot lists leave half of the opening states and two-thirds of the held items unstated. Our planner adds completions that agree with the film, raising the stated opening states to 45\%, 71\%, and 73\% and the stated held items to 22\%, 68\%, and 51\% for synopses, paragraphs, and shot lists in \cref{fig:planner_text_main}~(left). It keeps every requested event and 90--98\% of requested outcomes and invents at most 0.07 items per plan.

\paragraph{Cross-shot state.}
Code maintenance makes carrying a recorded change independent of the planner's strength. Qwen3.8-27B and the weaker Qwen3.5-9B each record about half of the outfit changes, and once a change is recorded, code states the post-change attire correctly in 66\% of later entries with either model. Asked instead to restate attire and held items in every shot, Qwen3.5-9B gets a recorded change right in only 35\%, so code significantly raises its post-change attire accuracy from 33.6\% to 47.7\%, while Qwen3.8-27B reaches 55.1\% with code and 52.7\% without, as App.~C details.

\paragraph{Effect on rendered stories.}
In the storyboard rows of \cref{tab:ablation_renderer}, our planner's storyboard matches the reference storyboard within 0.5 points on cut execution and characters, exceeds it by 9.3 points on Persist and 8.0 on Establish, and trails it by 5.6 points on objects. It exceeds the request by 5.4 points on characters, 16.0 on Establish, and 15.1 on Persist.

\paragraph{Written versus executed.}
Joining each state question with the per-shot text the renderer received, our planner states 91\% of the required changes, against 88\% for the reference storyboard and 75\% for the request, and the renderer executes 69\% of the changes our planner states, against 61\% and 62\% for the other two, as \cref{fig:planner_text_main}~(right) shows. On the 50 questions that both the reference and our planner state, 71\% are shown from our planner's text against 61\% from the reference, suggesting that its storyboard is easier to render.

\section{Conclusion}
\label{sec:conclusion}

We presented \method, which plans a whole story once and renders it window by window: a public language model writes a storyboard, code maintains its cross-shot state and compiles each shot's text, and the storyboard indexes which generated frames a window renderer trained on the same schema recalls. On \benchmark, \method attains the highest average score and the most consistent characters and keeps the most requested changes on return.

\label{last-body-page}

\section*{LLM Usage Statement}
Language models enter this work in two roles. As research components described in the paper, Qwen3.7-Flash and Qwen3.8-27B annotate the data, Qwen3.8-27B is also the planner, GPT-6.1-sol writes the benchmark questions, and Gemini 3.8 Flash and DeepSeek-V4.1-Flash serve as judges. As assistants, coding agents (Claude Code, OpenAI Codex, Cursor, and Google Antigravity) wrote and debugged code, implemented and ran the experiments the authors designed, computed results, generated the data plots, searched the literature for related work, and drafted and edited text, all of which the authors reviewed, revised, and approved; the authors drew the schematic figures. Claude and ChatGPT also gave reviewer-style feedback on drafts of the manuscript. The authors made the research decisions, designed the experiments, and take full responsibility for the content.

{
    \small
    \bibliographystyle{ieeenat_fullname}
    \bibliography{main}
}

\clearpage
\appendix
\section*{Appendix}
\renewcommand{\thefigure}{S\arabic{figure}}
\renewcommand{\thetable}{S\arabic{table}}
\renewcommand{\theequation}{S\arabic{equation}}
\setcounter{figure}{0}
\setcounter{table}{0}
\setcounter{equation}{0}
\renewcommand{\theHfigure}{appendix.\arabic{figure}}
\renewcommand{\theHtable}{appendix.\arabic{table}}
\renewcommand{\theHequation}{appendix.\arabic{equation}}
\makeatletter
\setlength{\@fptop}{0pt}
\setlength{\@dblfptop}{0pt}
\setlength{\@fpsep}{\floatsep}
\setlength{\@dblfpsep}{\dblfloatsep}
\makeatother

\noindent
This appendix gives what is needed to reproduce the method and the reported numbers:
\begin{itemize}
  \item \textbf{Section~\ref{supp:impl}}: renderer configuration (\cref{tab:supp_impl}), reference noise (\cref{fig:supp_ref_noise}), window settings (\cref{tab:supp_window}), and planner configuration.
  \item \textbf{Section~\ref{supp:planner}}: planner prompt and storyboard schema (\cref{tab:supp_prompt_planner}).
  \item \textbf{Section~\ref{supp:planner_text}}: text-level planner analyses (\cref{tab:supp_prompt_planner_judge,fig:planner_text,tab:supp_planner_state}) and the planner ablation on one development story (\cref{fig:planner_qual}).
  \item \textbf{Section~\ref{supp:data}}: data screening, annotation, spatial fields, and comparison with multi-shot datasets (\cref{tab:supp_prompt_annot,tab:datasets_train}).
  \item \textbf{Section~\ref{supp:benchmark}}: benchmark stories, questions, judges with their validation against human annotation, metrics, and comparison with multi-shot benchmarks (\cref{tab:supp_prompt_eval,tab:supp_questions,tab:supp_prompt_pair,tab:judge_agreement,tab:supp_ungated,tab:datasets_bench}).
  \item \textbf{Section~\ref{supp:baselines}}: baseline configurations and input conversion (\cref{tab:supp_configs,tab:supp_native_prompts}).
  \item \textbf{Section~\ref{supp:videos}}: index of the videos on the project page (\cref{tab:supp_videos}).
  \item \textbf{Section~\ref{supp:code}}: the code repository and its entry points (\cref{tab:supp_code}).
\end{itemize}

\section{Implementation Details}
\label{supp:impl}
\Cref{tab:supp_impl} lists the configuration of the renderer behind Table~1.

\begin{table*}[t]
  \centering
  \caption{\textbf{Renderer configuration.} Frame counts are planned at 16\,fps and rendered at 24\,fps.}
  \label{tab:supp_impl}
  \TableSetup
  \begin{tabularx}{\textwidth}{l L}
    \toprule
    \TableHead
    Item & Setting \\
    \midrule
    \PrimaryGroup{2}{Training} \\
    Model & Wan2.2-TI2V-5B; all 5.0B transformer parameters trained, text encoder and VAE frozen \\
    Objective & Rectified flow velocity target of Eq.~(3), with the MSE averaged over target latents and Wan's per-timestep weight; timesteps uniform with shift 5; reference latents excluded \\
    Video & $1280\times704$ at 24\,fps \\
    Windows & Complete source spans of at most 8\,s (128 frames at 16\,fps); a longer shot fills a window alone and shots over 10\,s are excluded; one, two, three, and four or more shots drawn at 15, 25, 30, and 30\%, half of the one-shot draws being shots of at least 5\,s with an action or a within-shot change; windows beyond the measured memory envelope of 8 GPUs (69 latent frames, 4,209 attended frame pairs) are excluded \\
    References & 1 to 10 frames from other usable shots of the same scene block, shots sharing an entity with the window first, one random frame per shot from the 16\,fps annotation grid, in shot order; each zoom-cropped to a random 65--100\% of its area at the original aspect ratio, the crop covering all registered characters in the frame and skipped when they do not fit (about 47\% cropped); temporal RoPE positions five latent frames apart, ending five before the reading shot's first latent frame; whole group dropped with probability 0.2; noise $\sigma \sim 0.2\,\mathcal{U}[0,1)$, applied as $(1-\sigma)x+\sigma\epsilon$ (\cref{fig:supp_ref_noise}); each frame carries an identity line listing the registered characters annotated as visible in it and the location of its source shot \\
    Text & Every registered character and object present in a shot is described; style sentence and composition sentence each dropped with probability 0.5 \\
    Optimization & AdamW ($\beta=(0.9, 0.999)$, weight decay 0.01), learning rate $10^{-5}$ with 300 warmup steps then constant, global batch 8 on 8 A100 GPUs, gradient clipping 1.0, BF16 autocast, ZeRO stage 2; 3,000 updates \\
    \midrule
    \PrimaryGroup{2}{Inference} \\
    Sampler & Official flow UniPC, 50 steps, shift 5, guidance 5 with Wan's default negative prompt on target shots only \\
    Windows & Consecutive whole shots packed under a cap of 7\,s (112 frames at 16\,fps) for the test stories and 8\,s (128 frames) for the development ablations, a longer shot filling a window alone, each shot text within 512 tokens; a planned count $F$ becomes $4\,\mathrm{round}((1.5F-1)/4)+1$ frames at 24\,fps; window $m$ uses seed $42+1009\,m$ \\
    References & Up to 10 frames of earlier generated shots, one per shot, chosen as below, without noise \\
    \bottomrule
  \end{tabularx}
\end{table*}

\begin{figure}[t]
  \centering
  \begin{subfigure}[t]{0.49\columnwidth}
    \centering
    \includegraphics[width=\linewidth]{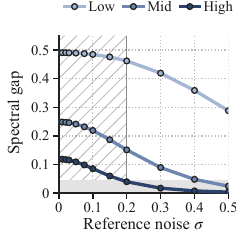}
  \end{subfigure}\hfill
  \begin{subfigure}[t]{0.49\columnwidth}
    \centering
    \includegraphics[width=\linewidth]{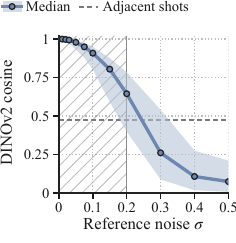}
  \end{subfigure}
  \caption{\textbf{Reference noise.} \textbf{(Left)}~Power gap between generated and real first latent frames, where Low, Mid, and High pool the lowest two, middle four, and highest two radial frequency bands and the gray band is the sampling spread of real frames. \textbf{(Right)}~DINOv2~\cite{oquab2024dinov2,liu2025dinov3} cosine between decoded noised and clean real references, median with 10--90\% range, against the median of two adjacent shots. Hatching marks the training range of $\sigma$.}
  \label{fig:supp_ref_noise}
\end{figure}

\paragraph{Reference noise and dropout.}
References are real frames in training but generated frames at inference. Noising them in training narrows this gap at the cost of their appearance, and \cref{fig:supp_ref_noise} measures both on 336 shots of 96 public multi-shot clips of two to five shots, held out from training and separate from the benchmark, comparing each real shot's first latent frame with that of a shot generated by an earlier checkpoint of our renderer, both decoded and re-encoded by the same VAE. Generated latents carry more power at every spatial frequency, 52\% more in the lowest band and 12\% more in the highest. By $\sigma=0.2$ the noise hides the high-frequency difference within the sampling spread of real frames, while a noised real reference still resembles its clean self more than two adjacent shots of one story resemble each other; from $\sigma=0.3$ it no longer does, and the low-frequency difference remains at every level. We therefore draw $\sigma$ uniformly from $[0, 0.2]$ in training and use clean references at inference. The whole reference group is dropped with probability 0.2, close to how often inference windows read no references: first windows make up 16\% of test windows and 20\% of development windows.

\paragraph{Reference recall at inference.}
Recall reads only the storyboard. Candidates are the shots generated before the window, and each offers one frame: its last frame if the shot records a change of attire or visible state, and its middle frame otherwise, including shots that change only a posture or a held item. For each target shot and each present registered character or object, the anchor view is the candidate showing that entity with the smallest key $(u, n, d, j)$ of Eq.~(1), compared in order. $u$ is 1 for a candidate that precedes the latest shot before the target that records a change of attire or visible state of the entity, and 0 otherwise. $n$ is the number of other registered characters in the candidate. $d$ applies to characters only: 0 for a close-up, medium close-up, or medium shot, 1 for an extreme close-up or medium wide shot, and 2 for a wide or extreme wide shot, with an unstated shot size counted as medium. $j$ is the shot index, so ties go to the earliest shot. Each target shot with a location also takes the candidate at that location with the fewest registered characters, then the widest shot size, then the earliest index. Picks are collected over the target shots in order, entities before the location, deduplicated by source shot, capped at ten, and placed in shot order. A recalled frame is read by the target shots that share a present entity with its source shot, and by all target shots when it shares none with the window. Each recalled frame carries the line \emph{``Reference still from another shot of this story; not the current shot. Visible identities: \dots\ Setting: \dots''}. At inference the identities are the registered characters that the storyboard places in the source shot and the setting is its location. In training they come from per-frame labels of the registered characters visible in the reference frame. The + first views row of the renderer ablation also gives each present entity the earliest shot showing it since its latest change of attire or visible state, once when that shot is already its anchor view, and gives the target shot's location its earliest shot as well.

\paragraph{Window length and packing.}
Training bounds window length only from above, so the inference cap is a free choice: with the 8\,s training cap, 36\% of multi-shot windows on the development stories last 7\,s or more, against 27\% of training windows. Caps of 7\,s and 6\,s change the windows of 23 and 29 development stories, and even packing, which keeps the 8\,s cap and the number of windows of greedy packing, changes 14; one shot per window, with the same recall, splits every multi-shot window. In each case the other stories have the same inputs and reuse the renders of \method. The 1-Wasserstein distance between the multi-shot window lengths and those of training is 0.91\,s at 6\,s, 0.58\,s at 7\,s, and 0.72\,s at 8\,s. Of the 308 consistency pairs on the development stories, 259 cross a window boundary under greedy packing. \Cref{tab:supp_window} renders these settings under the renderer ablation of Table~2.

\begin{table}[t]
  \centering
  \caption{\textbf{Window settings} on 30 development stories over three generation seeds, under the renderer ablation of Table~2.}
  \label{tab:supp_window}
  \TableSetup
  \begin{tabularx}{\columnwidth}{l *{6}{Y}}
    \toprule
    \TableHead
    Setting & Exec. & Char. & Obj. & Scene & Persist & Avg. \\
    \midrule
    \focusrow \method & 0.958 & 0.581 & 0.585 & 0.566 & 0.615 & 0.571 \\
    \PrimaryGroup{7}{Window cap} \\
    \quad 7\,s / window & 0.969 & 0.576 & 0.594 & 0.601 & 0.625 & 0.573 \\
    \quad 6\,s / window & 0.960 & 0.571 & 0.620 & 0.580 & 0.701 & 0.583 \\
    \PrimaryGroup{7}{Packing} \\
    \quad even packing & 0.958 & 0.578 & 0.538 & 0.549 & 0.635 & 0.570 \\
    \quad 1 shot / window & 0.967 & 0.578 & 0.483 & 0.649 & 0.609 & 0.579 \\
    \bottomrule
  \end{tabularx}
\end{table}

\paragraph{Copying.}
Copying is the share of shot pairs with a planned new camera setup, 387 per seed on the development stories including non-adjacent returns, that the camera-setup judge finds in the same setup or without a detected cut, averaged within a story and then over stories. Training without identity lines raises it from 3.56\% to 5.04\%, a difference of 1.47 points with the paired 95\% interval [0.29, 2.54].

\paragraph{Routing.}
Cut execution on the development stories scores, once per seed, the 241 of their 255 planned cuts between shots that the storyboard makes different. We group them by whether routing hides at least one of the window's recalled frames from the target shot; both rows recall the same frames. Where routing hides a frame, 264 cut--seed items in 26 stories, the renderer trained with routing executes 0.970 and the one trained without it 0.869, a difference of 10.1 points [3.6, 17.2]; 32 of the latter's 34 failures are judged the same setup and 2 are not cut. On the other 459 items in 30 stories they execute 0.953 and 0.939, a difference of 1.4 points [$-$2.7, 6.7].

\paragraph{Recall across windows.}
For character questions two or more windows apart on the development stories, 381 over three seeds in 27 stories, the recalled frames of the recent-shots row include the character in 348 (91\%) but come from the window of the character's first appearance in only 84 (22\%), against 228 (60\%) for storyboard recall (Figure~6).

\paragraph{CLIP retrieval.}
The retrieval control of the renderer ablation replaces this recall and keeps everything else. Each generated shot offers up to six candidate frames, a frame being dropped when its CLIP ViT-B/32 image embedding has a cosine of 0.95 or more with an earlier candidate of the same shot, with StoryMem's image preprocessing. The text of each shot in the window, truncated at 77 tokens, scores every candidate; each earlier shot is scored by its best frame, and the window takes the highest-scoring shots, one frame each, as many as the anchor views would recall.

\paragraph{Planner.}
Rows rendered from planner storyboards use Qwen3.8-27B served by vLLM 0.19 in BF16 with a 49,152-token context, thinking enabled at medium reasoning effort, the model card's thinking sampling (temperature 1.0, top-$p$ 0.95, top-$k$ 20, presence penalty 0, at most 32,768 tokens), and schema-constrained decoding with xgrammar (\cref{supp:planner}). The compiler sets the total frame count to the value nearest $16\,T$ that is a sum of $4n{+}1$ shot counts, distributes it in proportion to the proposed frame counts by greedy four-frame increments with every shot between 1 and 225 frames, and packs windows as above.

\section{Planner Prompt and Storyboard Schema}
\label{supp:planner}
The planner receives the system prompt of \cref{tab:supp_prompt_planner} and one user message: a JSON object with the request granularity (\texttt{summary}, \texttt{paragraph}, or \texttt{shot} for the synopses, paragraphs, and shot lists) and the request text, preceded by the production header \emph{``Production requirements: approximately $T$ seconds total; 16 fps; at most 225 frames per shot (14.0625 seconds).''} Decoding is constrained to a JSON schema with exactly the fields and values listed in the prompt, and every object forbids extra keys. When a shot-list plan has a different number of shots from the request, or a plan has too few shots to hold the requested duration, the same conversation receives one correction (part II) and the second answer is final.

\begin{PromptPlate}
  \caption{\textbf{Planner prompts.} Part I is the complete system prompt and part II the two structural corrections.}
  \label{tab:supp_prompt_planner}
\begin{tcolorbox}[prompt,title={Part I: Planner System Prompt},title after break={Part I: Planner System Prompt (continued)}]
You are the storyboard planner of a multi-shot video system. Turn the supplied tier and script into one JSON storyboard with characters, objects, places, and shots. Code then compiles the storyboard into one prompt per shot for a Wan video generator. Write concise English.\par
\textbf{What the code does after you}\par
$\bullet$ Copies each registered appearance and outfit into the shot prompts, so write them once in the registries.\par
$\bullet$ Allocates the requested total frame budget across shots and applies the generator's frame grid.\par
$\bullet$ Adds a bare present entry for an object that appears only in a holding array.\par
$\bullet$ Carries each character's current outfit, and a holding you leave unwritten, into later shots.\par
$\bullet$ Chooses reference frames from earlier shots by who is present.\par
It does not infer missing actions or state changes.\par
\textbf{Steps}\par
1. Register every character (including groups and animals), object, and place the request mentions as Character N, Object N, Location N. Keep supplied IDs; otherwise number each category from 1.\par
2. Lay out the shots. Read the frame rate and maximum shot duration from Production requirements. For tier=shot, keep every given shot in its order; do not merge, add, or remove shots. For tier=summary or paragraph, use at least the total seconds divided by the maximum shot duration, rounded up, and enough shots to show every requested event.\par
3. For each shot, write each entity's initial visible properties (present), the event with its visible progression (action), and the result at its end (changes).\par
4. Propose each shot's frame\_count. Add the opening composition only when the request states or implies how the opening frame is laid out.\par
\textbf{Three kinds of values}\par
$\bullet$ Stated by the request: copy registered appearance and outfit, places, camera instructions, and visual medium faithfully. Preserve every requested event, its performer, order, and required result. The action wording may change under the action rules below; do not weaken, omit, or replace a requested event.\par
$\bullet$ Implied by the request or by earlier shots (who is present, initial visible posture and state, possession, end-of-shot changes): write every one, consistent with earlier shots.\par
$\bullet$ Neither stated nor implied (details a coarse request leaves open, pacing): choose one concrete, plausible value that fits the request, the camera, and the action, and keep it the same across shots of the same setup. Never add plot events or subjects the request does not ask for, and do not write mood or atmosphere prose.\par
\textbf{Fields}\par
characters: name ``Character N''; kind person \textbar{} animal \textbar{} group; appearance and outfit: when the request describes the character, copy its words exactly, including any pose, accessory, or expression it mentions; otherwise write the stable look and the default clothing as lowercase noun phrases (``an adult courier'', ``a blue jacket''). Leave outfit out when the request gives none for a group or an animal. visibility clear \textbar{} partial \textbar{} distant, only when the request says how the character is seen.\par
objects, places: name ``Object N'' or ``Location N''; appearance: copy the request's description exactly when it gives one; otherwise write a lowercase noun phrase (``a small brown parcel'', ``a sunlit workshop'').\par
shots, one entry per shot in order:\par
$\bullet$ shot: 1, 2, 3, ...\par
$\bullet$ place: a registered location.\par
$\bullet$ picture\_type: live\_action \textbar{} animation\_2d \textbar{} animation\_3d\_or\_game \textbar{} screen\_or\_slides, only when the request states the medium.\par
$\bullet$ camera, only the parts the request gives: shot\_size extreme close-up shot \textbar{} close-up shot \textbar{} medium close-up shot \textbar{} medium shot \textbar{} medium wide shot \textbar{} wide shot \textbar{} extreme wide shot; angle eye-level shot \textbar{} high angle shot \textbar{} low angle shot \textbar{} overhead shot \textbar{} over-the-shoulder shot \textbar{} dutch angle shot \textbar{} aerial shot; movement static \textbar{} pan \textbar{} tilt up \textbar{} tilt down \textbar{} push in \textbar{} pull out \textbar{} tracking \textbar{} handheld.\par
$\bullet$ composition: center composition \textbar{} balanced composition \textbar{} symmetrical composition \textbar{} left-heavy composition \textbar{} right-heavy composition, the words Wan's own prompt expander uses. Write it only when the request states or implies how the opening frame is laid out (a person framed alone on one side, two people placed symmetrically); otherwise leave it out.\par
$\bullet$ present: every participating character and object, each as it looks when it first becomes visible in this shot, with posture, state, holding (an array of registered entity names), and outfit only when it differs from the registered one. An entity that enters later need not be visible at the shot's start; preserve the entry or exit in action. Holding means physically holding or carrying the entity, including a person or animal, with hands, mouth, or body. Being near, pointing at, entering, sitting in, or riding an entity does not mean holding it; describe those interactions in action or posture.\par
$\bullet$ action: concise, ordered sentences describing this shot's event. Start with the performing registered subject (Character N or Object N as appropriate) and one short distinguishing feature from its registry or current outfit; do not repeat the full appearance. For an object-only or static shot, describe the requested subject or visible state without inventing a character or movement. Keep every requested action and its performer, and keep where the request places people (on the left, behind, in the background) in these sentences. When useful, infer a short, physically plausible progression that makes the requested action visible, without adding a new plot event, subject, or stronger ending. Inferred intermediate steps are a proposed execution, not observations of an unseen source video. Use then or finally only when the actions are sequential; preserve simultaneous actions as simultaneous.\par
$\bullet$ changes: the entity, what (posture \textbar{} state \textbar{} outfit \textbar{} holding), before (equal to the opening), and after (the visible ending).\par
$\bullet$ ending\_sides: characters who end the shot on another side of the frame, with left \textbar{} center \textbar{} right, only when the request says so.\par
$\bullet$ frame\_count: relative pacing in frames, not seconds; multiply seconds by the frame rate in Production requirements. Longer actions need more time than reactions; keep requested timing; do not solve the total.\par
\textbf{Writing rules}\par
1. Use only registered names, exactly as registered, in every field and sentence. A change belongs to an existing entity; never register it as a new one.\par
2. The opening never contains the shot's result: a person who stands up starts seated; a cake being cut starts uncut.\par
3. State and posture values are lowercase complements that can follow ``is'', with no subject, copula, or final period: ``empty-handed'', ``with her hair loose'', ``in a plain white T-shirt'', ``with a fresh cut on her cheek''.\par
4. Describe a result by what is visible, not by what is gone: never ``without'', ``removed'', ``no longer'', ``not wearing'', ``not holding'', or ``no''. An action may still name the item (``takes off the headband''), and an item that stays visible keeps its name.\par
5. A transfer releases the object from the giver and adds it to the recipient; later shots keep the possession until another release. When a holding array ends empty, also write the visible state (``empty-handed'', or ``with its mouth empty'' for an animal).\par
6. Write an outfit change once, in changes, with the full new outfit as after; put uncovered hair or skin in state.\par
7. Do not replay a finished event to restate its result; carry the result into later openings.\par
8. Expand only what helps execute the requested action. Do not invent movements for a static shot or pad the action into a fixed number of steps. Match the detail to the available time; when the request fixes timing, keep it and avoid an overlong action chain.\par
\textbf{Example}\par
Input:\par
\begin{lstlisting}[style=PromptJSON]
{"tier":"shot","script":"Production requirements: approximately 12 seconds total; 16 fps; at most 225 frames per shot (14.0625 seconds).\n\nUser request:\nCreate a multi-shot video using the following 3 shots in order.\n\nSUBJECTS\n- Character 1 (person): an adult courier, wearing a blue jacket\n- Character 2 (person): an adult woman, wearing a grey coat\n- Object 1 (object): a small brown parcel\nLOCATIONS\n- Location 1: a sunlit workshop\nSHOTS\ns01: Character 1 holds Object 1 and greets Character 2.\ns02: Character 1 hands Object 1 to Character 2.\ns03: Character 2 carries Object 1 to the window while Character 1 waits by the door."}
\end{lstlisting}
Output:\par
\begin{lstlisting}[style=PromptJSON]
{"characters":[{"name":"Character 1","kind":"person","appearance":"an adult courier","outfit":"a blue jacket"},{"name":"Character 2","kind":"person","appearance":"an adult woman","outfit":"a grey coat"}],"objects":[{"name":"Object 1","appearance":"a small brown parcel"}],"places":[{"name":"Location 1","appearance":"a sunlit workshop"}],"shots":[{"shot":1,"place":"Location 1","present":[{"name":"Character 1","posture":"standing","holding":["Object 1"]},{"name":"Character 2","posture":"standing","holding":[]}],"action":"Character 1, the courier in the blue jacket, greets Character 2, the woman in the grey coat, while holding Object 1.","changes":[],"frame_count":64},{"shot":2,"place":"Location 1","present":[{"name":"Character 1","posture":"standing","holding":["Object 1"]},{"name":"Character 2","posture":"standing","holding":[]}],"action":"Character 1, the courier in the blue jacket, extends Object 1 toward Character 2, the woman in the grey coat. Character 2 takes Object 1, and Character 1 lets go.","changes":[{"name":"Character 1","what":"holding","before":["Object 1"],"after":[]},{"name":"Character 1","what":"state","after":"empty-handed"},{"name":"Character 2","what":"holding","before":[],"after":["Object 1"]}],"frame_count":64},{"shot":3,"place":"Location 1","present":[{"name":"Character 1","posture":"standing by the doorway","state":"empty-handed","holding":[]},{"name":"Character 2","posture":"standing","holding":["Object 1"]}],"action":"Character 2, the woman in the grey coat, carries Object 1 to the window while Character 1, the courier in the blue jacket, waits by the door.","changes":[{"name":"Character 2","what":"posture","before":"standing","after":"standing at the window"}],"frame_count":64}]}
\end{lstlisting}
The parcel changes hands once and remains with Character 2. Extending, taking, and letting go are a plausible inferred execution of the requested handover, not observations of an unseen source video. The compiler keeps this expanded action, adds the referenced parcel's bare present entry, and resolves the total frame budget. Do not copy the example's characters or story into a different request.\par
Return only the JSON object.\par
\end{tcolorbox}
\begin{tcolorbox}[prompt,title={Part II: Structural Corrections}]
\RoleTag{USER} The request lists exactly \{required\} shots, but your storyboard has \{planned\}. Return the complete storyboard again with exactly \{required\} shots, one per listed shot, in the order of the request.\par
\RoleTag{USER} Your storyboard has \{planned\} shots, but \{target\_seconds\} seconds at 16 fps is \{frames\} frames and one shot holds at most 225 frames, so it needs at least \{minimum\} shots. Return the complete storyboard again with at least \{minimum\} shots, keeping everything the request states.
\end{tcolorbox}
\end{PromptPlate}

\section{Text-Level Planner Analyses}
\label{supp:planner_text}
All text-level analyses use the frozen planner prompt and two sampling seeds unless stated, DeepSeek-V4.1-Flash as the text judge, and 95\% intervals resampled by source film, both seeds of a story drawn together; the 30 development stories come from 23 films. The 40 unseen stories are blocks from 40 films outside the benchmark, the development set, and the samples used while writing the prompt. Re-judging identical plans changes the stated outcomes and the contradicted checkpoints by about two and the fabricated items by about four per 30 plans.

Request fidelity uses the text-only judge of \cref{tab:supp_prompt_planner_judge}. Checkpoints are fixed before any plan is read: for development shot lists they are the action and state questions of the benchmark bank, and for development synopses and paragraphs Gemini 3.8 Flash writes them once from each request. For the 40 unseen stories, the judge model writes the checkpoints of every granularity once from the request, without shot anchors. The judge reads the request without the production header, the compiled per-shot text of the plan (or its raw JSON when it does not compile), and the checkpoints, at high reasoning effort with JSON output.

\begin{PromptPlate}
  \caption{\textbf{Planner text judge prompt.} The user message lists the request, the plan, and one line per checkpoint.}
  \label{tab:supp_prompt_planner_judge}
\begin{tcolorbox}[prompt,title={Planner Text Judge Prompt},title after break={Planner Text Judge Prompt (continued)}]
\RoleTag{SYSTEM} You check a written storyboard plan against the user request it was made from. The request alone decides what is required; the plan may add details the request leaves open. The user message gives the request, the plan, and checkpoints that were written from the request before any plan was read.\par
For each checkpoint decide what the plan writes:\par
$\bullet$ stated: the plan writes this event or result (paraphrase is fine; the plan uses ids such as Character 1 for people the request describes).\par
$\bullet$ contradicted: the plan writes something incompatible with it (another doer or receiver, the opposite direction, the result undone, the event replaced by a different one).\par
$\bullet$ absent: the plan does not write it.\par
Give the plan shot number where it is stated or contradicted, or null when absent. For event checkpoints give roles: correct when the doer and the receiver match the request, wrong when either is another person or object; use not\_applicable for results and for absent checkpoints.\par
Then list contradictions: anything the plan writes that conflicts with the request (an identity, appearance, place, action, order or outcome the request fixes). Do not list details the request leaves open, and do not repeat checkpoints judged contradicted.\par
Output one json object with exactly the keys ``checkpoints'' and ``contradictions'': one checkpoints entry per checkpoint, its id copied exactly as written inside the square brackets, status one of stated \textbar{} contradicted \textbar{} absent, roles one of correct \textbar{} wrong \textbar{} not\_applicable.\par
EXAMPLE JSON OUTPUT (for checkpoints [e01] and [e01r]):\par
\begin{lstlisting}[style=PromptJSON]
{"checkpoints": [{"id": "e01", "status": "stated", "shot": 2, "roles": "correct"}, {"id": "e01r", "status": "absent", "shot": null, "roles": "not_applicable"}], "contradictions": [{"shot": 3, "plan": "Character 2 sits at the desk.", "request": "The woman stands by the window."}]}
\end{lstlisting}
\RoleTag{USER} \texttt{USER REQUEST} \{request\}\par
\texttt{PLAN} (the per-shot text a video renderer reads, compiled from the storyboard) \{plan text\}\par
\texttt{CHECKPOINTS} \texttt{[id] event (request shot n, during the shot): question} \textbar{} \texttt{[id] state (request shot n, by the end of the shot): question} \textbar{} \dots
\end{tcolorbox}
\end{PromptPlate}

\paragraph{Information in the text.}
The left panel of Figure~7 compares the request and the compiled per-shot text against the opening states and held items of the reference storyboard, whose entries (1,468 for the development stories, 1,930 for the unseen stories) the text judge marks stated, contradicted, or absent. The planner states 34\%, 33\%, and 48\% of the missing opening states for synopses, paragraphs, and shot lists, lifting the totals to 45\%, 71\%, and 73\%, and 16\%, 25\%, and 29\% of the missing held items (totals 22\%, 68\%, and 51\%). A completion counts only when it agrees with the film. What the request asks for is kept: against checkpoints written from the request alone, the text states every requested event at every granularity and 90\%, 98\%, and 96\% of the requested outcomes, with at most 0.01 contradicted and 0.07 fabricated items per plan.

\begin{figure}[t]
  \centering
  \includegraphics[width=\columnwidth]{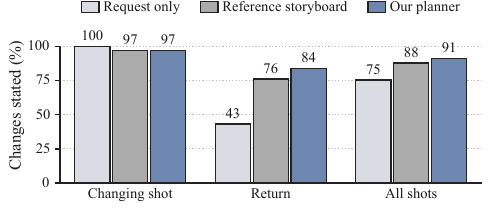}
  \caption{\textbf{State changes in the text} of the planner ablation: the share of state questions whose change the per-shot text states in the changing shot, when the character returns, and over all its shots.}
  \label{fig:planner_text}
\end{figure}

\paragraph{Cross-shot state.}
The change sample has 100 blocks from films outside the benchmark, drawn so that 60 contain an outfit change. Both forms use the frozen prompt with three field descriptions reordered to follow the schema. The model-maintained variant replaces two sentences: code ``does not carry anything between shots, so write each character's current outfit and holding in every shot where it appears'', and after an outfit change the model writes the new outfit ``again as the outfit of every later shot's present entry where it is worn''. Code then carries nothing for it. The pair is run with Qwen3.8-27B as configured above and with Qwen3.5-9B without thinking, sampled as the model card recommends for that mode. The judge reads each checkpoint against that shot's view only: the compiled text for ours, and the registry with that shot's JSON entry for the model-maintained form (\cref{tab:supp_planner_state}).

\begin{figure*}[t]
  \centering
  \includegraphics[width=\textwidth]{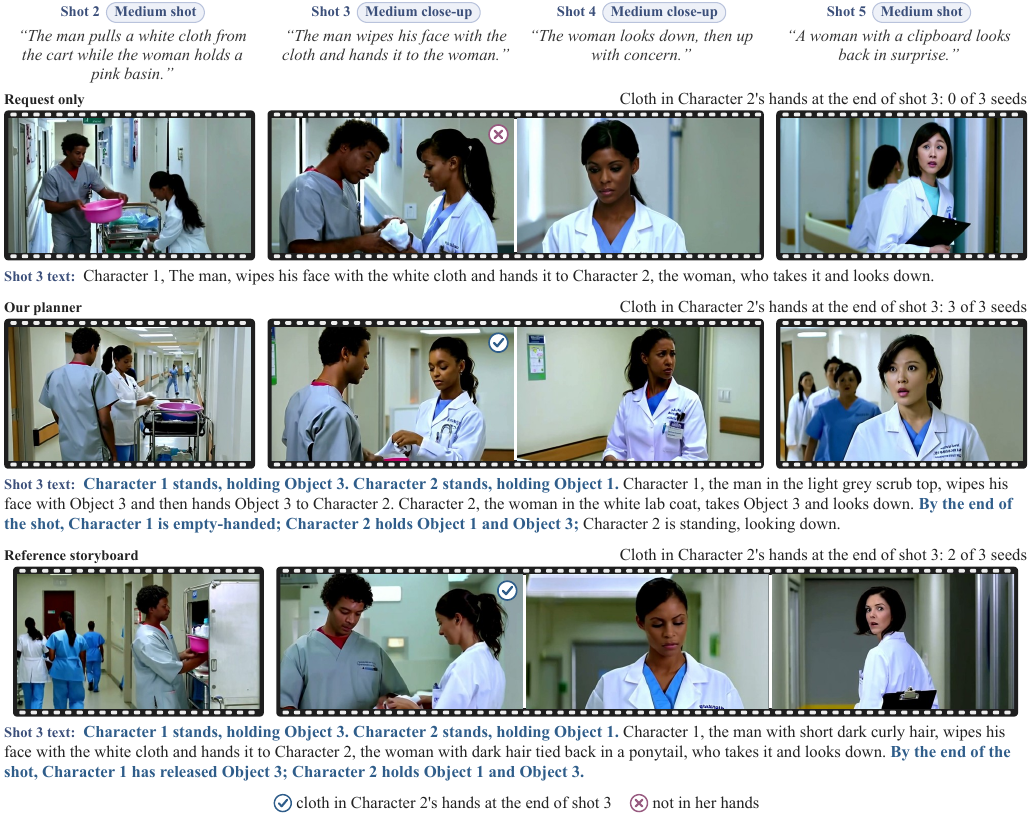}
  \caption{\textbf{Planner on one development story}, seed 44. Each film strip is one generation pass of the same renderer, which reads the storyboard of its row. Shot 3 is shown at its end and the other shots at their middle frame. Under each row is the shot-3 text the renderer read, with the opening and closing states in bold; Object 3 is the white cloth, and the marks give the judge's answer on whether Character 2 holds it at the end of shot 3.}
  \label{fig:planner_qual}
\end{figure*}

\paragraph{Written versus executed.}
We join each state question of the storyboard rows of Table~2 with the per-shot text the renderer received, read by the text-only judge of \cref{tab:supp_prompt_planner_judge}. Compiled directly, the request states every change in the shot that makes it but restates only 43\% of them when the character returns, against 84\% for our planner (\cref{fig:planner_text}). In the story of \cref{fig:planner_qual}, all three texts state a hand-over, and it is rendered in 0, 3, and 2 of three seeds from the request, our planner, and the reference, the latter two also writing who holds what as the shot opens and ends.

\begin{table}[t]
  \centering
  \caption{\textbf{Cross-shot state} stated correctly in later shots (\%), change sample, two seeds, for each planner model. $^\dagger$ marks a significant difference from code. The indented rows split the outfit entries after a change by whether the plan records that change.}
  \label{tab:supp_planner_state}
  \TableSetup
  \begin{tabularx}{\columnwidth}{l *{2}{Y}}
    \toprule
    \TableHead
    Entries & Code & Model \\
    \midrule
    \PrimaryGroup{3}{Qwen3.8-27B, thinking at medium effort} \\
    Held item (454) & 59.0 & 57.0 \\
    Outfit (1,730) & 85.9 & 82.7$^\dagger$ \\
    Outfit after change (256) & 55.1 & 52.7 \\
    \quad change recorded & 52.7 & 53.9 \\
    \quad correct if recorded & 65.9 & 65.2 \\
    \quad correct if not & 43.0 & 39.5 \\
    \PrimaryGroup{3}{Qwen3.5-9B, no thinking} \\
    Held item (454) & 56.6 & 55.1 \\
    Outfit (1,730) & 83.9 & 81.7 \\
    Outfit after change (256) & 47.7 & 33.6$^\dagger$ \\
    \quad change recorded & 52.0 & 53.1 \\
    \quad correct if recorded & 66.2 & 35.3 \\
    \quad correct if not & 27.9 & 31.7 \\
    \bottomrule
  \end{tabularx}
\end{table}

\section{Film Storyboard Data}
\label{supp:data}

\paragraph{Shot screening and blocks.}
TransNetV2 boundaries are refined and two guard frames are trimmed at each end of a shot. A shot then qualifies if it lasts at least 1\,s and at most 14.25\,s (228 frames at 16\,fps), comes from a source of at least 16\,fps with an active area of at least $832\times480$ after border cropping, has a mean Rec.~709 luminance in $[10, 210]$, and contains no interior cut: no TransNetV2 frame score of 0.25 or more and no PySceneDetect content score of 27 or more away from its ends. Scenes with burned-in subtitles are dropped. Consecutive qualified shots of one scene with touching source intervals and a LAION aesthetic score of at least 4.0 form runs, and a run longer than twelve shots is split into near-equal blocks of at most twelve. After annotation, a shot enters the training pool only if its LAION aesthetic score, averaged over its first, middle, and last frames, is at least 4.5, its DOVER technical score is at least $-0.149$ (one standard deviation below the LSVQ test mean), its annotation passes verification, and its compiled text fits 512 umT5 tokens. A failing shot stays in its block as a breakpoint that no training window crosses. Blocks are split by film, with 2\% of films held out.

\paragraph{Shot-first against window-first screening.}
The 28.1\% gain in usable shots, and an 18.9\% gain in adjacent shot pairs, compare two screenings inside the same annotated blocks. Shot-first screening cuts maximal runs of passing shots into non-overlapping four-shot windows; window-first screening cuts each block into non-overlapping four-shot windows from its first shot and drops a window when any of its shots fails. Shot-first screening keeps 290,267 shots against 226,648, and 166,550 adjacent pairs, consecutive shots of one scene within 14.25\,s, against 140,120.

\paragraph{Annotation.}
Qwen3.7-Flash, with thinking enabled and JSON output, annotates each block in one request: a proxy video of the block at $832\times480$ and 2\,fps, the prompt of \cref{tab:supp_prompt_annot}, and the time range of every shot. Verification rejects a block whose story fields, registry, or shot numbering are malformed or whose shots reference unregistered entities, and deletes single invalid entries inside a shot before re-checking it; a shot that still fails becomes a breakpoint, and rejected or unanswered blocks are re-sent. A within-shot change is recorded in 22.6\% of shots.

\begin{PromptPlate}
  \caption{\textbf{Film storyboard annotation prompt.} The shot time ranges follow as lines such as \texttt{Shot 1: 0.0\ttdash{}3.5 s}.}
  \label{tab:supp_prompt_annot}
\begin{tcolorbox}[prompt,title={Annotation Prompt},title after break={Annotation Prompt (continued)}]
You will watch a short video and write structured notes about it. The notes are used to train a video-generation model. The video is one scene made of one or more shots; a shot is one continuous piece of footage between two cuts. The time range of every shot is listed at the end.\par
Work in three steps, in this order: first list who and what is in the scene, then describe every shot, then tell the whole scene once.\par
Write only what you can see. If you cannot tell something, leave it out; do not guess. Do not use subtitles, titles, or what you know about the film or its actors. Do not say where things are in the frame (left, right, foreground, background) or which way people face; that is measured separately. Use short, plain English. Answer with one JSON object with the keys ``characters'', ``objects'', ``places'', ``shots'' and ``story'', and nothing else.\par
\textbf{Step 1. characters, objects, places: who and what is in the scene}\par
Number them in the order they first appear. Every entry of characters is labelled ``Character 1'', ``Character 2'', ... whether it is a person, an animal or a group; every object is ``Object 1'', ``Object 2'', ...; every place is ``Location 1'', ``Location 2'', ... These labels are fixed codes, not names you choose: a baboon is ``Character 3'', never ``Baboon 1''; a savanna is ``Location 1'', never ``Savanna''. What it is goes into appearance. Write the label in the ``name'' field and use it everywhere else. The same person, animal, thing or place keeps its label in every shot. If you are not sure that two appearances are the same one, give them different labels.\par
$\bullet$ characters: the people and animals the scene is about, one entry each: those who act, speak or are clearly shown. Everyone else goes into one entry per group, with kind ``group'': passers-by, an audience, people in the background, a herd, a row of soldiers. Describe a group by what it is, such as ``guests at the other tables''.\par
\hspace*{1em}kind: ``person'', ``animal'' or ``group''.\par
\hspace*{1em}visibility: ``clear'' (a visible face or a distinctive whole figure), ``partial'' (only hands, feet, a back or other parts) or ``distant'' (always too small or too far away to tell apart).\par
\hspace*{1em}appearance: what stays the same for the whole video, such as hair, build, skin tone, age, beard and face. Leave out what they wear, pose, expression and anything temporary.\par
\hspace*{1em}outfit: what they wear in this scene, such as ``a navy sweater and jeans''. Leave it out if you cannot see what they wear, and for animals that wear nothing.\par
$\bullet$ objects: every thing you can clearly see and identify: things people hold or use, furniture, vehicles, doors, windows and so on. Clothes that someone is wearing belong to that character's outfit, not here.\par
\hspace*{1em}appearance: what kind of thing it is, with its colour, material and shape. Leave out its momentary state, such as open, full or broken.\par
$\bullet$ places: each room or site. Two shots show the same place if they show the same room or site, even from a different camera position.\par
\hspace*{1em}appearance: what the place looks like.\par
\textbf{Step 2. shots: one entry per shot, in order}\par
$\bullet$ shot: its number, starting at 1.\par
$\bullet$ picture\_type: what kind of picture this shot is. One of:\par
\hspace*{1em}``live\_action'': filmed with a camera in the real world.\par
\hspace*{1em}``animation\_2d'': drawn or flat cartoon animation.\par
\hspace*{1em}``animation\_3d\_or\_game'': computer-generated 3D animation, video-game footage, stop-motion or puppet animation, or a rendered visualisation.\par
\hspace*{1em}``screen\_or\_slides'': a screen recording, a still image, or a page of text or graphics rather than moving footage.\par
$\bullet$ on\_screen\_text: text or graphics laid over the picture in this shot. Text that belongs to the filmed scene, such as a shop sign or a phone screen, does not count. One of:\par
\hspace*{1em}``none'': nothing is laid over the picture.\par
\hspace*{1em}``corner\_logo\_or\_watermark'': only a small logo, watermark, score bug or short label, in a corner or for a moment.\par
\hspace*{1em}``burned\_subtitles'': subtitle or caption lines are burned into the picture for most of the shot.\par
\hspace*{1em}``large\_text\_or\_picture\_in\_picture'': large text blocks, a title card, a decorative frame, a split screen or picture-in-picture cover a substantial part of the picture for most of the shot.\par
$\bullet$ place: a location label. Leave place and camera out for a shot that is not footage of a scene, such as a title card or a page of text.\par
$\bullet$ camera: leave out any part you cannot tell.\par
\hspace*{1em}shot\_size: ``extreme close-up shot'', ``close-up shot'', ``medium close-up shot'', ``medium shot'', ``medium wide shot'', ``wide shot'' or ``extreme wide shot''.\par
\hspace*{1em}angle: ``eye-level shot'', ``high angle shot'', ``low angle shot'', ``overhead shot'', ``over-the-shoulder shot'', ``dutch angle shot'' or ``aerial shot''.\par
\hspace*{1em}movement: ``static'', ``pan'', ``tilt up'', ``tilt down'', ``push in'', ``pull out'', ``tracking'' or ``handheld''.\par
$\bullet$ present: every character and object visible in this shot, each as it looks when it first appears in the shot. A hand, a back or another part of a character counts: list that character.\par
\hspace*{1em}name\par
\hspace*{1em}posture (characters): body position and where they are, such as ``sitting at the dining table'' or ``standing in the doorway''.\par
\hspace*{1em}state: anything temporary you can see, such as ``sweating'', ``a bandaged arm'', ``open'', ``full of water'' or ``lying on the table''.\par
\hspace*{1em}holding (characters): names of the objects in their hands.\par
\hspace*{1em}outfit (characters): only when what they wear in this shot differs from their outfit in step 1, for example a coat taken off or a hat put on; then write everything they wear now, and keep writing it in every later shot until they wear their step-1 outfit again. Leave it out otherwise.\par
$\bullet$ activity: what the people or animals in this shot do. One of:\par
\hspace*{1em}``action\_or\_state\_change'': someone visibly does something with their body or with an object, or something visibly changes: picks up, puts down, puts on, opens, pours, cuts, hands over, sits down, stands up, walks somewhere, enters, leaves, hugs, hits, dances, plays a sport, cooks.\par
\hspace*{1em}``talking\_only'': people only talk, sing, pose, look around or gesture; nothing else happens.\par
\hspace*{1em}``no\_human\_activity'': nobody acts: scenery, empty rooms, objects or vehicles on their own.\par
$\bullet$ action: one sentence saying what happens in this shot. Refer to characters, objects and places by their names and do not repeat their appearance.\par
$\bullet$ changes: what becomes different during this shot. One entry per change: name (a character or object listed in present); what (``outfit'', ``posture'', ``state'' or ``holding''); before; after. For ``holding'', before and after are lists of object names. Write [] if nothing changes.\par
\textbf{Step 3. story: tell the whole scene once}\par
$\bullet$ paragraph: a few sentences telling what happens from the first shot to the last, as one continuous story. Tell only things you wrote down in step 2. Do not mention shots, cuts or the camera. Refer to each person by a few words, such as ``the grey-haired woman'', not by their name.\par
$\bullet$ summary: one sentence saying who does what and where. Refer to people the same way as in paragraph.\par
\textbf{Example with two shots}\par
\begin{lstlisting}[style=PromptJSON]
{"characters": [{"name": "Character 1", "kind": "person", "visibility": "clear", "appearance": "an elderly woman with short grey hair and a slim build", "outfit": "a navy sweater"}, {"name": "Character 2", "kind": "person", "visibility": "clear", "appearance": "a tall person with long brown hair", "outfit": "a long grey coat"}], "objects": [{"name": "Object 1", "appearance": "a white ceramic mug"}, {"name": "Object 2", "appearance": "a brown leather suitcase"}, {"name": "Object 3", "appearance": "a wooden dining table"}, {"name": "Object 4", "appearance": "a glass back door"}], "places": [{"name": "Location 1", "appearance": "a small kitchen with wooden cabinets"}], "shots": [{"shot": 1, "picture_type": "live_action", "on_screen_text": "none", "place": "Location 1", "camera": {"shot_size": "medium shot", "angle": "eye-level shot", "movement": "static"}, "present": [{"name": "Character 1", "posture": "sitting at the dining table"}, {"name": "Character 2", "posture": "walking in through the back door", "state": "wet hair", "holding": ["Object 2"]}, {"name": "Object 1", "state": "full of tea, standing on the table"}, {"name": "Object 2"}, {"name": "Object 3"}, {"name": "Object 4", "state": "open"}], "activity": "action_or_state_change", "action": "Character 1 lifts Object 1 from Object 3 and drinks while Character 2 comes in through Object 4 carrying Object 2.", "changes": [{"name": "Character 1", "what": "holding", "before": [], "after": ["Object 1"]}]}, {"shot": 2, "picture_type": "live_action", "on_screen_text": "none", "place": "Location 1", "camera": {"shot_size": "close-up shot", "angle": "eye-level shot", "movement": "static"}, "present": [{"name": "Character 1", "posture": "sitting at the dining table", "holding": ["Object 1"]}, {"name": "Character 2", "posture": "standing by the dining table", "outfit": "a white shirt and dark trousers"}, {"name": "Object 1", "state": "empty"}, {"name": "Object 3"}], "activity": "talking_only", "action": "Character 1 looks up and speaks to Character 2, who has taken off the coat.", "changes": []}], "story": {"paragraph": "An elderly woman sits at a wooden table in a small kitchen and lifts a mug of tea to drink. A tall person with wet hair comes in through the glass back door carrying a leather suitcase. Later the tall person, now without the coat, stands by the table while the woman looks up and talks.", "summary": "An elderly woman drinks tea in a kitchen as a tall person comes in from outside with a suitcase."}}
\end{lstlisting}
\textbf{Example with one shot}\par
\begin{lstlisting}[style=PromptJSON]
{"characters": [{"name": "Character 1", "kind": "person", "visibility": "clear", "appearance": "a young man with curly dark hair and a lean build", "outfit": "a grey hoodie and black shorts"}, {"name": "Character 2", "kind": "animal", "visibility": "clear", "appearance": "a small brown dog"}, {"name": "Character 3", "kind": "group", "visibility": "distant", "appearance": "joggers on the path"}], "objects": [{"name": "Object 1", "appearance": "a green wooden bench"}], "places": [{"name": "Location 1", "appearance": "a tree-lined park path"}], "shots": [{"shot": 1, "picture_type": "live_action", "on_screen_text": "corner_logo_or_watermark", "place": "Location 1", "camera": {"shot_size": "medium wide shot", "angle": "eye-level shot", "movement": "static"}, "present": [{"name": "Character 1", "posture": "sitting on the bench, bent over his shoe"}, {"name": "Character 2", "posture": "lying under the bench"}, {"name": "Character 3", "posture": "running along the path"}, {"name": "Object 1"}], "activity": "action_or_state_change", "action": "Character 1 bends forward on Object 1, ties the laces of one shoe, then sits up and looks down the path as Character 3 runs past.", "changes": [{"name": "Character 1", "what": "posture", "before": "bent forward over his shoe", "after": "sitting upright on the bench"}]}], "story": {"paragraph": "A young man sits on a green bench beside a park path with a small brown dog lying under it. He bends forward and ties the laces of one shoe, then sits up and looks down the path as a group of joggers runs past.", "summary": "A young man ties his shoelaces on a park bench while joggers pass behind him."}}
\end{lstlisting}
\end{tcolorbox}
\end{PromptPlate}

\begin{table*}[t]
  \centering
  \caption{\textbf{Multi-shot training data in detail.} Size counts clips, episodes for DramaBoard, and blocks for ours. Present per shot lists the registered entities in each shot, and Camera gives each shot's size, angle, or movement. Per shot gives each entity's posture, clothing, or held object, and before/after the initial and final values of a within-shot change. $^\ast$ marks artifacts listed on the project page.}
  \label{tab:datasets_train}
  \TableSetup
  \setlength{\tabcolsep}{2.3pt}
  \begin{tabular}{l>{\hspace{4pt}}l!{\color{grid}\vrule}c!{\color{grid}\vrule}cccc!{\color{grid}\vrule}cc!{\color{grid}\vrule}ccc!{\color{grid}\vrule}c!{\color{grid}\vrule}cc}
    \toprule
    \TableHead
    & & Shots & \multicolumn{4}{c!{\color{grid}\vrule}}{Entity registry} & \multicolumn{2}{c!{\color{grid}\vrule}}{State} & \multicolumn{3}{c!{\color{grid}\vrule}}{Grounding} & & \multicolumn{2}{c}{Release} \\
    \TableHead
    & & per & & & & Present & Per & Before/ & & & & & & \\
    \TableHead
    \multicolumn{1}{c}{\multirow{-3}{*}{Dataset}} & \multicolumn{1}{c!{\color{grid}\vrule}}{\multirow{-3}{*}{Size}} & sample & \multirow{-2}{*}{Char.} & \multirow{-2}{*}{Object} & \multirow{-2}{*}{Place} & per shot & shot & after & \multirow{-2}{*}{Box} & \multirow{-2}{*}{Mask} & \multirow{-2}{*}{Depth} & \multicolumn{1}{c!{\color{grid}\vrule}}{\multirow{-3}{*}{Camera}} & \multirow{-2}{*}{Labels} & \multirow{-2}{*}{Videos} \\
    \midrule
    HoloCine~\cite{meng2026holocine} & 400K & $\le$13 & \ckH & \ckN & \ckN & \ckH & \ckH & \ckN & \ckN & \ckN & \ckN & \ckH & No & No \\
    MultiShotMaster~\cite{wang2026multishotmaster} & 235K & 1--5 & \ckY & \ckY & \ckN & \ckY & \ckH & \ckN & \ckY & \ckH & \ckN & \ckH & No & No \\
    CineDance-1M~\cite{chen2026cinedance} & 1.02M & 24.2 & \ckY & \ckN & \ckY & \ckY & \ckH & \ckN & \ckN & \ckN & \ckN & \ckY & Partial & Partial \\
    DramaBoard~\cite{zhou2026dramadirector} & 2.8K & 29 & \ckY & \ckN & \ckN & \ckY & \ckY & \ckN & \ckN & \ckN & \ckH & \ckY & Public & No \\
    MSAV-Script-100K~\cite{zhao2026ssrs} & 100K & 5.3 & \ckY & \ckY & \ckH & \ckH & \ckH & \ckH & \ckN & \ckN & \ckN & \ckY & Public & Public \\
    Mise-en-Video~\cite{zou2026miseenvideo} & 160K & 2--6 & \ckH & \ckN & \ckN & \ckH & \ckH & \ckN & \ckY & \ckH & \ckH & \ckH & No & No \\
    \focusrow Our data & 75K & 4.7 & \ckY & \ckY & \ckY & \ckY & \ckY & \ckY & \ckY & \ckY & \ckY & \ckY & Public$^\ast$ & Public$^\ast$ \\
    \bottomrule
  \end{tabular}
  \\[2pt]
  \begin{minipage}{\textwidth}\footnotesize
    \MarkLegend
  \end{minipage}
\end{table*}

\paragraph{Spatial fields.}
Spatial fields are measured on every fourth frame of the 16\,fps grid. Qwen3.8-27B sees the first, middle, and last of these frames and returns, for each present character and object other than groups, a box at the earliest frame where it is clearly visible. SAM~3.1 propagates a mask from that box through all frames, and VGGT-$\Omega$ estimates depth, so each instance has the median depth inside its mask. The first and last frames where an entity's mask covers at least 0.2\% of the frame are its two anchors. From the characters visible at the opening, the compiler derives the composition from the area-weighted mean of their horizontal box centers: left-heavy below 0.4 and right-heavy above 0.6; otherwise center for one character or a spread below one third of the width, symmetrical for two characters placed symmetrically within 0.1 with an area ratio of at least 0.5, and balanced in all other cases. Neighboring characters whose centers differ by at least 0.15 of the width are related by \texttt{left\_of}, and neighbors in the depth ranking whose median log-depth ratio exceeds 0.1 by \texttt{in\_front\_of}. A character changes screen side when its horizontal third differs between its two anchors and its center moves by at least 0.15.

\paragraph{Comparison with multi-shot datasets.}
\Cref{tab:datasets_train} details the dataset comparison that the related work summarizes in one sentence. It covers the training data of the two multi-shot generators we compare against, HoloCine~\cite{meng2026holocine} and MultiShotMaster~\cite{wang2026multishotmaster}, the CineDance-1M pool, one of the three sources of our blocks~\cite{chen2026cinedance}, and the closest prior record for each column, namely DramaBoard, the storyboard set of DramaDirector~\cite{zhou2026dramadirector}, for typed per-shot states, MSAV-Script-100K~\cite{zhao2026ssrs} for within-shot changes, and Mise-en-Video~\cite{zou2026miseenvideo} for entity grounding. Marks follow each paper and its released files. An identity type is provided when later shots reuse its identifiers, a state is provided when it is a typed field rather than a sentence, and a box or mask counts when it is part of the annotation rather than an intermediate of the pipeline. A partial mark means the field exists only as free text, covers only part of the entities, frames, or samples, appears only in a prompt format, or is only partly released.

\section{\benchmark Protocol}
\label{supp:benchmark}

\paragraph{Stories.}
Candidates come from the held-out split, 916 blocks from 189 films that no training block uses. Adjacent blocks of one film are rejoined, split at shots that fail screening, and cut greedily into non-overlapping windows of six to twelve shots and at most 60 seconds. A window is kept only if it contains a within-shot state change, a character recurring across non-adjacent shots, and a dynamic action shot, which gives 122 stories (1,117 shots, 67 films): 30 for development and 92 for testing (832 shots), with no shared or overlapping source clip. Every method renders each test story once under seed 42 without retries.

\paragraph{Questions.}
For each story, GPT-6.1-sol reads the request with its annotation and lists the entities of each shot, the actions that act on an object or another person or move the whole body, and the lasting changes they cause (\cref{tab:supp_prompt_eval}, part I). A deterministic program turns this list into binary questions that name entities by registry identifier (\cref{tab:supp_questions}). Character, object, and scene questions compare an entity's first appearance with its second, with every return after an off-screen gap, and with its last. An Establish question asks whether a change holds at the end of the shot that requests it, and a Persist question whether it still holds when the entity reappears. We label each state question by reading its shot's request: an Establish question is scored only when the request writes the change and the bank holds a question for the action that makes it, and it passes only when the judge confirms both. Object questions are kept only where the object description reaches both compared shots in every baseline's actual prompt. The test set has 1,012 scored questions (662 character, 65 object, 194 scene, 44 Establish, and 47 Persist) and the development set 349 (225, 21, 62, 16, and 25).

\FloatBarrier
\begin{PromptPlate}
  \caption{\textbf{Question-writer and whole-video judge prompts.}}
  \label{tab:supp_prompt_eval}
\begin{tcolorbox}[prompt,title={Part I: Checklist Question-Writer Prompt},title after break={Part I: Checklist Question-Writer Prompt (continued)}]
\RoleTag{SYSTEM} You prepare checklist questions for judging a generated multi-shot video against the user request below. A structured annotation of the same story is attached: its IDs match the request and help you keep people and objects apart. The request alone decides what is required.\par
\textbf{1. Names:} For each registered ID, give the name the request uses for it, such as \texttt{"Subject 2"}, \texttt{"Location 1"}, or the short phrase \texttt{"the birthday woman"}.\par
\textbf{2. Shots:} For each shot of the request, list the IDs of the people and objects the request says are in that shot.\par
\textbf{3. Actions:} List every action the request describes that acts on an object or on another person, or moves the whole body such as standing up, sitting down, walking away, or falling. Steps that serve one goal are one action: \emph{``raises the glass, takes a sip and lowers it''} is drinking. Leave out facial expressions, gaze, speaking, listening, camera, and background. Write each action as a short yes/no question about that one action alone, such as \texttt{"Does Subject 1 drink the red wine?"}, and list the participating IDs. If the action changes something that stays changed afterwards, also write one yes/no question about that new state, say whose state it is, and whether the change is to \texttt{"appearance"} or \texttt{"holding\_or\_place"}; otherwise give \texttt{null}.
\end{tcolorbox}
\begin{tcolorbox}[prompt,title={Part II: Whole-Video Judge Prompt for Action and State Questions},title after break={Part II: Whole-Video Judge Prompt (continued)}]
\RoleTag{SYSTEM} \textbf{Task.} You check visible events and states in a generated multi-shot video. You will receive a registry identifying characters, objects and locations, each numbered shot's time range, and a numbered list of yes/no questions. Each question asks about one thing that should be visible in its given time range. The registry identifies entities; it does not establish that any requested action or state appears in the video.\par
\textbf{How to answer each question.}\par
$\bullet$ \texttt{yes}: you can clearly see it in the video within the given time range.\par
$\bullet$ \texttt{no}: the required action or state is absent, or the visible evidence is insufficient to establish it.\par
$\bullet$ For an event asking for an action or transition, observe the process within the time range; an end result alone is insufficient. If the person or object already has the target pose or state at the start and remains static throughout, with no required transition, answer no. A state question can be yes when that state is visibly present at the requested time without a transition.\par
$\bullet$ The registry only tells you which character, object or location a question is about. Find it by the role the question gives it in this time range (who performs the action, which object is used, where it happens). Do not answer no because the person, object or place looks different from the registry description or from other shots; identity across shots is checked separately. Answer no for a missing entity only when nothing in the time range plays that role. A character shown from behind or in profile still counts.\par
$\bullet$ An action near the frame edge counts as yes when the action trajectory and interaction are clearly visible, even if part of an object is cropped (for example pouring into a container partly below the bottom edge). If the required interaction itself cannot be seen, answer no.\par
$\bullet$ Judge every question independently and use only evidence inside its specified time range. Treat ranges as start-inclusive and end-exclusive: the first frame of the next shot is not evidence for the preceding shot.\par
$\bullet$ For both yes and no, describe the observed initial and final states and the observed transition or state features. State when something is not visible; never invent a trajectory from an expected outcome.\par
\textbf{Output.} For each question, give the time (\texttt{m:ss.d}) where you looked and a brief reason in this order: [Initial $\rightarrow$ Final] the observed state at the start and at the end of the time range; [Transition] the specific observed action trajectory or state features, including any missing transition. Then answer yes or no.
\RoleTag{USER} \texttt{[Complete Generated Video, 2 fps]} \par
\texttt{\#\# Registry: SUBJECTS - \{registered character or object\} | ... LOCATIONS - \{registered location\} | ...} \par
\texttt{\#\# Shots in this video: Shot 1: 0:00.0\ttdash{}0:03.2 | Shot 2: 0:03.2\ttdash{}0:06.5 | ...} \par
\texttt{\#\# Questions: 1. (end of 0:00.0\ttdash{}0:03.2) \{state\_establishment\_question\} | 2. (0:06.5\ttdash{}0:09.8) \{state\_revisit\_question\} | ...} \par
Based on the video, registry and time ranges above, answer every question.
\end{tcolorbox}
\end{PromptPlate}

\begin{table}[t]
  \centering
  \caption{\textbf{Example questions} from two test stories, one per scored dimension.}
  \label{tab:supp_questions}
  \TableSetup
  \begin{tabularx}{\columnwidth}{l L}
    \toprule
    \TableHead
    Dimension & Question \\
    \midrule
    Character & Is Character 1 at shot 3 the same person as at shot 1? \\
    Object & Is Object 5 at shot 10 the same object as at shot 8? \\
    Scene & Are shot 2 and shot 1 in the same place (Location 1)? \\
    Establish & End of shot 3: Is Object 3 open? Passes only with ``Does Character 5 open Object 3?'' in shot 3. \\
    Persist & Shot 4: Is Object 3 open? \\
    \bottomrule
  \end{tabularx}
\end{table}

\paragraph{Judges.}
Gemini 3.8 Flash answers every question at low reasoning, and unclear evidence counts as no. Action and state questions are answered in one call on the complete video at 2\,fps and medium media resolution, given the entity registry and the detected shot intervals but not the per-shot script; a shot's interval starts from the frame counts the method rendered, each planned boundary moved to the nearest detected cut within 0.5\,s, and a boundary with no cut that close stays at its planned time, so shot $k$ of a question always maps to the $k$-th planned shot; the judge finds an entity by the role the question gives it, so identity is scored only by the consistency questions (\cref{tab:supp_prompt_eval}, part II). Consistency questions are answered in one request per compared pair of shots from the first, middle, and last frames of each shot (\cref{tab:supp_prompt_pair}, part III).

\begin{PromptPlate}
  \caption{\textbf{Shot-pair and camera-setup judge prompts.}}
  \label{tab:supp_prompt_pair}
\begin{tcolorbox}[prompt,title={Part III: Shot-Pair Judge Prompt for Consistency Questions},title after break={Part III: Shot-Pair Judge Prompt (continued)}]
\RoleTag{SYSTEM} \textbf{Task.} You check a generated multi-shot video for consistency between two of its shots. You will see the first, middle and last frame of each shot, and a numbered list of questions. Each question asks whether one character, object or location in the later shot is the same individual as in the earlier shot.\par
\textbf{Rules.}\par
$\bullet$ Use only the frames. Each frame is labeled with its shot and position before the image.\par
$\bullet$ Compare the individuals in the two shots with each other. The registry description only tells you which character, object or location a question is about; do not judge how well a shot matches the description.\par
$\bullet$ Same character: the same person, judged by the face, hairstyle and build. Matching clothing, accessories or role alone is not evidence of the same person; a different face in the same outfit is a different person. Clothing may change between shots and does not by itself make two people different. If the face cannot be seen in one of the shots, the same hairstyle, build and clothing together count as the same person.\par
$\bullet$ Same object: the same specific item, judged by its shape, color, texture and distinctive details; another object of the same kind is not the same object.\par
$\bullet$ Same location: the same specific place, judged by fixed background features visible in both shots, such as walls, windows, furniture layout or landmarks. A different viewing angle of the same place is still the same place. The same type of place is not enough.\par
$\bullet$ A change of rendering style between live action, 2D animation and 3D animation means a different individual.\par
$\bullet$ If the character, object or location cannot be found in either shot, or the frames do not show enough to tell (for example it is out of frame, hidden, or the frame is filled by something else), the answer is no.\par
\textbf{Output.} For each question, first compare the listed features (same, different, or not visible), then answer yes or no. Use the question numbers exactly as given.
\RoleTag{USER} \texttt{\#\# Registry: - \{each asked character, object, or location\}} \par
\texttt{Shot 2 (earlier shot), first frame [image] | Shot 2 (earlier shot), middle frame [image] | ... | Shot 5 (later shot), last frame [image]} \par
\texttt{\#\# Questions: 1. [character] Is Character 1 at Shot 5 the same person as at Shot 2? | 2. [location] Are Shot 5 and Shot 2 in the same place (Location 1)? | ...} \par
Based on the frames above, answer each question in the group for its type.\par
\textbf{Response schema.} Each character answer compares face, hair, build, and clothing as \texttt{same}, \texttt{different}, or \texttt{not visible} before \texttt{same\_person}; each object answer compares shape, color and texture, and details before \texttt{same\_object}; each location answer compares layout, landmarks, and color and mood before \texttt{same\_place}.
\end{tcolorbox}
\begin{tcolorbox}[prompt,title={Part IV: Camera-Setup Judge Prompt},title after break={Part IV: Camera-Setup Judge Prompt (continued)}]
\RoleTag{SYSTEM} You check whether a cut in a generated multi-shot video moves to a new camera setup. You will see four frames: the first and last frame of Shot A, then the first and last frame of Shot B. Shot B directly follows Shot A in the video, or is a later shot that is compared with Shot A.\par
A camera setup is where the camera stands, which way it points, and how wide or close it frames the place. Judge it only by the large fixed structures of the place (walls, windows, doors, building fronts, the horizon): where they sit in the frame and how large they appear.\par
$\bullet$ \texttt{new}: the camera clearly moved, turned or zoomed, so these structures appear in different places or at a clearly different size, or a different part of the place is visible, or it is clearly a different place.\par
$\bullet$ \texttt{same}: these structures sit at about the same places and sizes. People moving closer or further away, entering, leaving or being replaced by someone else, and changes of small objects, lighting or surface detail do not make a new setup. If the camera position, viewing angle, and background framing remain the same, and the only difference is that a character moves closer or shifts position within the same fixed frame, answer same. This remains same when the moving foreground character covers part of the fixed background. A slight pan or small reframing is the same setup.\par
$\bullet$ \texttt{unclear}: only when a frame shows nothing usable, for example it is completely black. If a person or object fills the frame, still decide new or same from what you can see. A character appearing larger is not by itself evidence of a new setup; use the fixed background framing to distinguish camera reframing from foreground movement.\par
Use only the four frames. First describe each shot in one sentence: where the camera is and where the large fixed structures sit in the frame. Then answer.
\RoleTag{USER} \texttt{Shot A, first frame [image] | Shot A, last frame [image] | Shot B, first frame [image] | Shot B, last frame [image]} \par
Based on the four frames above, describe Shot A and Shot B, then answer: was Shot B filmed from a new camera setup?\par
\textbf{Response schema.} \texttt{shot\_a} and \texttt{shot\_b}, one sentence each, then \texttt{setup} as \texttt{new}, \texttt{same}, or \texttt{unclear}.
\end{tcolorbox}
\end{PromptPlate}

\paragraph{Validation against human annotation.}
Two annotators and a reviewer labeled questions of all seven baselines without seeing the method or the judge's answer, gated as the judge is gated. Each dimension has 13 to 17 shared questions, sampled to balance the judge's yes and no answers. For cut execution the judge answers yes on 86 of its 115 items. The annotators labeled 560 items, 94 of them in common, on which they agree at $\kappa=0.64$ before review. The reviewer resolved the 116 items on which the judge and an annotator, or the two annotators, disagree, labeling each from the frames before seeing any label, and found three of them unclear, which leaves 557. \Cref{tab:judge_agreement} reports Cohen's $\kappa$ and balanced accuracy between the judge and these reviewed labels, and the agreement of each annotator's own blind labels with them, $\kappa=0.853$ [0.80, 0.90] pooled over 532 annotator--item pairs. These gate with each annotator's own camera-setup labels. The 95\% intervals of $\kappa$, from resampling whole shared questions, are [0.46, 0.73] for characters, [0.48, 0.78] for objects, [0.38, 0.77] for scenes, [0.60, 0.90] for Establish, [0.62, 0.89] for Persist, [0.51, 0.88] for cut execution, and [0.64, 0.77] overall. The overall balanced accuracy has the interval [0.82, 0.88]. Scene agreement is lowest because after a shot--reverse-shot cut the later shot shows a side of the place the earlier one never showed.

\begin{table}[t]
  \centering
  \caption{\textbf{Agreement between the judge and reviewed human labels.} Bal.\ acc.\ averages the judge's accuracy on items the labels pass and on those they fail, and Annot.\ $\kappa$ compares each annotator's blind labels with the reviewed labels, pooled over 532 annotator--item pairs.}
  \label{tab:judge_agreement}
  \TableSetup
  \begin{tabularx}{\columnwidth}{l *{4}{Y}}
    \toprule
    \TableHead
    Dimension & Items & Judge $\kappa$ & Bal.\ acc. & Annot.\ $\kappa$ \\
    \midrule
    Execution & 115 & 0.724 & 0.906 & 0.903 \\
    Character & 87 & 0.609 & 0.809 & 0.845 \\
    Object & 91 & 0.641 & 0.836 & 0.719 \\
    Scene & 88 & 0.592 & 0.796 & 0.761 \\
    Establish & 87 & 0.770 & 0.885 & 0.810 \\
    Persist & 89 & 0.775 & 0.888 & 0.946 \\
    \midrule
    All & 557 & 0.704 & 0.851 & 0.853 \\
    \bottomrule
  \end{tabularx}
\end{table}

\begin{table*}[t]
  \centering
  \caption{\textbf{What the gate changes, and ViCLIP against the judge} on the 92 test stories. Gated checks is the share of consistency checks behind a failed gate. ViCLIP pairs shots showing the same character or all shots.}
  \label{tab:supp_ungated}
  \TableSetup
  \begin{tabularx}{\textwidth}{l!{\color{grid}\vrule}YY!{\color{grid}\vrule}YY!{\color{grid}\vrule}Y!{\color{grid}\vrule}YY}
    \toprule
    \TableHead
    & & & \multicolumn{2}{c!{\color{grid}\vrule}}{Inter-shot Consistency} & Judge & \multicolumn{2}{c}{ViCLIP} \\
    \TableHead
    Method & Execution & Gated checks & With gate & No gate & Character & Same character & All pairs \\
    \midrule
    Source films & -- & -- & -- & -- & -- & 0.665 & 0.598 \\
    \PrimaryGroup{8}{Per-shot generation} \\
    Wan2.1, per shot & 0.994 & 0.000 & 0.233 & 0.233 & 0.219 & 0.658 & 0.583 \\
    Wan2.2, per shot & 0.996 & 0.000 & 0.271 & 0.271 & 0.227 & 0.654 & 0.578 \\
    StoryMem & 0.719 & 0.157 & 0.590 & 0.765 & 0.631 & 0.749 & 0.674 \\
    StoryDiffusion & 0.825 & 0.092 & 0.270 & 0.307 & 0.300 & 0.653 & 0.561 \\
    IC-LoRA & 0.961 & 0.003 & 0.236 & 0.236 & 0.168 & 0.603 & 0.578 \\
    \PrimaryGroup{8}{Joint multi-shot generation} \\
    MultiShotMaster & 0.990 & 0.008 & 0.458 & 0.463 & 0.538 & 0.616 & 0.542 \\
    HoloCine & 0.967 & 0.009 & 0.538 & 0.542 & 0.517 & 0.629 & 0.587 \\
    \focusrow \method & 0.974 & 0.007 & 0.593 & 0.612 & 0.651 & 0.664 & 0.595 \\
    \bottomrule
  \end{tabularx}
\end{table*}

\paragraph{Cut execution and the new-setup gate.}
A planned cut between shots that the storyboard makes different, by place, shot size, angle, or cast, is executed when TransNetV2 detects a cut within 0.5\,s of it and the later shot opens a new camera setup. A cast change means that the two shots share no character. Of the 740 planned cuts of the test set, 674 are of this kind: the shot size changes by one step or more at 461 of them, the cast at 440, the angle at 120, and the place at 18. At the other 66 the storyboard keeps all four, so these cuts enter Cut F1 but not Execution, and no check across them is gated. Two shots whose middle frames, as $96\times54$ gray images, correlate at 0.9 or more are the same setup; every other pair goes to the camera-setup judge with the first and last frames of both shots and without the request (\cref{tab:supp_prompt_pair}, part IV), and an unclear answer counts as the same setup. The same decision gates consistency and Persist questions: across a pair that the storyboard makes different, $y(q)=0$ unless a cut is detected between the two shots and the later one opens a new setup. The test set has 1,097 such pairs, the 674 planned cuts among them.

\paragraph{Readings without the gate and with ViCLIP.}
To see what the gate changes, we sent every consistency check behind a failed gate on the test set to the same pair judge, 254 checks in 196 shot pairs over six methods; the per-shot Wan baselines have none. \Cref{tab:supp_ungated} reports, for each method, the share of its 921 consistency checks behind a failed gate and the consistency aspect with and without the gate. StoryMem's failed gates hide 145 checks, of which the judge finds 124 consistent. Without the gate StoryMem would lead every method, the next being \method by 15.3 points [9.8, 20.1]; with it the two differ by 0.3 [$-$6.1, 6.3]. Every other method changes by at most 3.7 points.

The table also gives the ViCLIP inter-shot similarity that HoloCine reports, over shot pairs showing the same character, and that StoryMem reports on ST-Bench, over all shot pairs, next to the judge's character score. Both ViCLIP readings take eight frames per shot, cut at the frame counts each method was asked to render, and the source films are cut at their annotated shot boundaries. ViCLIP disagrees with the judge: over same-character pairs it rates StoryMem 8.4 points above the source films and places the two per-shot Wan baselines, whose characters the judge finds consistent in 22--23\% of checks, within 1.1 points of the films, and across the eight methods its Kendall $\tau$ with the judge's character score is 0.21. The human labels side with the judge. On the items of \cref{tab:judge_agreement}, ViCLIP similarity reaches an AUC of 0.63, 0.70, and 0.49 for characters, objects, and scenes, and even a threshold tuned on these labels gives $\kappa$ of 0.27, 0.31, and 0.10, against 0.61, 0.64, and 0.59 for the judge. A video that repeats its first shot in every shot scores 1.0 on either ViCLIP reading. Under the gate, every planned new setup correlates at 1.0 with the shot before, so the video executes no cut, and even if every ungated check passed, its consistency would be at most 0.24 (characters 0.33, objects 0.25, scenes 0.14).

\begin{table*}[t]
  \centering
  \caption{\textbf{Multi-shot benchmarks in detail.} Columns follow \cref{tab:datasets_train} without grounding, which no benchmark here labels. Size counts stories, episodes for EntityBench, clips for CineBench, and scripts for CAST-Bench. A dash marks a script with no video, and $^\ast$ marks artifacts listed on the project page.}
  \label{tab:datasets_bench}
  \TableSetup
  \setlength{\tabcolsep}{7pt}
  \begin{tabular}{l>{\hspace{4pt}}l!{\color{grid}\vrule}c!{\color{grid}\vrule}cccc!{\color{grid}\vrule}cc!{\color{grid}\vrule}c!{\color{grid}\vrule}cc}
    \toprule
    \TableHead
    & & Shots & \multicolumn{4}{c!{\color{grid}\vrule}}{Entity registry} & \multicolumn{2}{c!{\color{grid}\vrule}}{State} & & \multicolumn{2}{c}{Release} \\
    \TableHead
    & & per & & & & Present & Per & Before/ & & & \\
    \TableHead
    \multicolumn{1}{c}{\multirow{-3}{*}{Benchmark}} & \multicolumn{1}{c!{\color{grid}\vrule}}{\multirow{-3}{*}{Size}} & sample & \multirow{-2}{*}{Char.} & \multirow{-2}{*}{Object} & \multirow{-2}{*}{Place} & per shot & shot & after & \multicolumn{1}{c!{\color{grid}\vrule}}{\multirow{-3}{*}{Camera}} & \multirow{-2}{*}{Labels} & \multirow{-2}{*}{Videos} \\
    \midrule
    MSVBench~\cite{shi2026msvbench} & 20 & 13.8 & \ckY & \ckN & \ckY & \ckY & \ckH & \ckN & \ckY & Public & -- \\
    EntityBench~\cite{he2026entitybench} & 140 & 17.8 & \ckY & \ckY & \ckY & \ckY & \ckH & \ckN & \ckH & Public & No \\
    CineBench~\cite{chen2026cinedance} & 1K & 2--20 & \ckY & \ckN & \ckY & \ckY & \ckH & \ckN & \ckY & No & No \\
    StoryEngine~\cite{wang2026storyengine} & 60 & 10 & \ckH & \ckH & \ckY & \ckH & \ckH & \ckY & \ckN & No & -- \\
    CAST-Bench~\cite{liu2026castbench} & 900 & 2.8 & \ckY & \ckY & \ckH & \ckY & \ckH & \ckH & \ckY & Public & -- \\
    \focusrow \benchmark & 122 & 9.2 & \ckY & \ckY & \ckY & \ckY & \ckY & \ckY & \ckY & Public$^\ast$ & Public$^\ast$ \\
    \bottomrule
  \end{tabular}
  \\[2pt]
  \begin{minipage}{\textwidth}\footnotesize
    \MarkLegend
  \end{minipage}
\end{table*}

\begin{table*}[t]
  \centering
  \caption{\textbf{Baseline configurations and input conversion.} Frame counts planned at 16\,fps are kept, except for official Wan2.2, which rounds $1.5\times$ the planned count to the nearest $4n{+}1$.}
  \label{tab:supp_configs}
  \TableSetup
  \begin{tabularx}{\textwidth}{l P{6.1cm}@{\hspace{7pt}}L}
    \toprule
    \TableHead
    Method & Configuration & Input conversion \\
    \midrule
    Wan2.1-T2V-1.3B & $832\times480$, 16\,fps, UniPC 50 steps, CFG 6, shift 8 & One shot per call with its compiled caption, no prompt extension; shots concatenated. \\
    Wan2.2-TI2V-5B & $1280\times704$, 24\,fps, UniPC 50 steps, CFG 5, shift 5 & As above. \\
    MultiShotMaster & $832\times480$, 50 steps, CFG 5, shift 5; up to 5 shots and 19.3\,s (308 frames) per segment; saved at 16\,fps & \texttt{Story:} global caption and \texttt{Now:} shot caption, with \texttt{Subject N} tags for characters, objects, and places. \\
    HoloCine & $832\times480$, 50 steps, CFG 5, shift 5; up to 15.1\,s (241 frames) and 512 tokens per segment; saved at 16\,fps & Global caption, per-shot captions joined by \texttt{[shot cut]}, \texttt{[characterN]} tags, and cut indices rounded to $4n{+}1$ as the model requires. \\
    StoryMem & $832\times480$, 16\,fps, UniPC 40 steps, CFG 3.5, shift 12 for the first shot and 4 after; memory of up to 10 keyframes (first 3 fixed, newest 7) & Each shot rendered at its planned length from its caption; every shot boundary updates the memory. \\
    StoryDiffusion + Wan2.2-I2V & SDXL, DDIM 35 steps, CFG 5; Wan2.2-I2V-A14B, UniPC 40 steps, CFG 3.5, shift 5 & SDXL draws each shot's keyframe from its opening state with consistent characters; the I2V model animates it with the full caption, including the end-state clause. \\
    IC-LoRA + Wan2.2-I2V & FLUX.1-dev with the film-storyboard LoRA, 20 steps, guidance 3.5; I2V stage as above & Three-panel $832\times1440$ grids cut into $832\times480$ keyframes, later grids keeping an earlier panel by inpainting; I2V stage as above. \\
    \bottomrule
  \end{tabularx}
\end{table*}

\paragraph{Window boundary and elapsed time.}
For HoloCine and MultiShotMaster, we split the character checks by whether the two shots fall in the same generation window of that method and by the time between the starts of the two shots, pooling the checks. Within a window they pass 71, 80, 83, and 83\% for HoloCine and 66, 62, 73, and 74\% for MultiShotMaster at under 3\,s, 3--6\,s, 6--10\,s, and at least 10\,s apart. Pooled over 3--10\,s, they pass 81.1\% [71.7, 89.0] and 66.0\% [56.2, 78.1] within a window and 27.8\% [21.5, 36.8] and 29.5\% [19.4, 42.9] across windows, gaps of 53.2 [40.0, 63.7] and 36.5 [21.3, 51.8] points, with intervals from resampling the 45 source films that have such checks. Split instead by HoloCine's windows and averaged per story, the character checks of \method lead HoloCine's by 22.2 points [12.2, 33.3] across those windows over 83 stories and differ by $-$0.9 [$-$9.5, 8.0] within them over 86 stories.

\paragraph{Metrics.}
With $y(q)\in\{0,1\}$ for each question or planned cut $q$ in the set $\mathcal{Q}_{i,d}$ of story $i$ and dimension $d$,
\begin{equation*}
  S_{i,d} = \frac{1}{|\mathcal{Q}_{i,d}|} \sum_{q \in \mathcal{Q}_{i,d}} y(q),
\end{equation*}
and each dimension is reported as the mean of $S_{i,d}$ over the stories that have items of it. For Cut F1, let $\{\tau_k\}_{k=1}^{K-1}$ be the requested cut times of a story with $K$ shots and $\{\hat{\tau}_{j}\}_{j=1}^{\hat{K}-1}$ the cut times detected by TransNetV2. Requested and detected cuts are matched one to one within 0.125\,s, two frames at 16\,fps, closest pairs first, giving the matching $\mathcal{M}$ and
\begin{equation*}
  \mathrm{F1} = \frac{2\,|\mathcal{M}|}{(K - 1) + (\hat{K} - 1)},
\end{equation*}
with $\mathrm{F1}=1$ when neither side has a cut, averaged over stories. Text alignment is the ViCLIP cosine similarity between eight frames of each shot, cut at the frame counts the method was asked to render, and that shot's line in the shot-list request, averaged over the shots of a story; ViCLIP truncates text at 32 tokens. The average of Table~1 is the mean of four aspects, each the mean of its applicable metrics: transition control (Execution and Cut F1), inter-shot consistency (Character, Object, Scene), state continuity (Establish, Persist), and text alignment. Differences between methods and their 95\% intervals follow the definition in the experimental setup.

\paragraph{Comparison with multi-shot benchmarks.}
\Cref{tab:datasets_bench} applies the same marks to multi-shot benchmarks that score entities or states across shots, namely MSVBench~\cite{shi2026msvbench}, EntityBench~\cite{he2026entitybench}, CineBench~\cite{chen2026cinedance}, StoryEngine~\cite{wang2026storyengine}, and CAST-Bench~\cite{liu2026castbench}. Our benchmark pairs registry identifiers with per-shot states and within-shot changes, which the Establish and Persist questions above require.

\section{Baseline Configurations}
\label{supp:baselines}
Every baseline renders the same reference storyboard, converted into the input format of its official examples and run with its official configuration (\cref{tab:supp_configs}). Every format carries the same storyboard fields: the registry appearances, the place, the shot size, angle, and camera movement, the posture, outfit, state, and held items of each entity present, the action, the end-of-shot changes, and the shot length. Joint generators read the appearances once in a global caption and refer to entities by their native tags, while per-shot methods repeat them in every shot. Official Wan2.1 and Wan2.2 leave out the default eye-level angle and static camera, and the keyframe prompts leave the action and the end state to the I2V caption. \Cref{tab:supp_native_prompts} shows one shot of a test story in each format. A story longer than a model's window is split at whole shots, never inside a shot, and segment $m$ uses seed $42 + 1009\,m$; StoryMem instead seeds shot $k$ with $42 + k$, and the SDXL keyframes of StoryDiffusion all use seed 42. Official Wan2.2 and \method produce $1280\times704$ videos at 24\,fps and all other methods $832\times480$ at 16\,fps, and every video reaches the judges as the original file. Due to encoder token limits, SDXL truncates 346 of the 832 keyframe prompts of StoryDiffusion at 77 CLIP tokens, and FLUX truncates the CLIP branch of IC-LoRA while its T5 branch reads the full grid caption.

\begin{PromptPlate}
  \caption{\textbf{Native inputs of one shot.} Shot 8 of a test story, in which a woman draws a handgun, as each baseline received it. The storyboard entry and its registry come first.}
  \label{tab:supp_native_prompts}
\begin{tcolorbox}[prompt,title={Shot 8 in the storyboard and in each native format},title after break={Shot 8 in each native format (continued)}]
\textbf{Reference storyboard.} \texttt{place}: Location 2. \texttt{camera}: medium shot, low angle shot, static. \texttt{present}: Character 3, standing on the escalator, holding nothing, and Object 2. \texttt{action}: The woman raises a handgun and points it forward. \texttt{changes}: Character 3 holding, from nothing to Object 2. \texttt{frame\_count}: 29. Registry: Character 3 is a woman with short brown hair, outfit a zebra-patterned coat and dark pants. Object 2 is a small black handgun. Location 2 is a lobby or hallway with an escalator.\par
\textbf{Wan2.1 and Wan2.2, per shot.} Medium shot, low angle shot. In a lobby or hallway with an escalator, Character 3, a woman with short brown hair, is standing on the escalator, wearing a zebra-patterned coat and dark pants. The woman raises a handgun and points it forward. By the end of the shot, Character 3 holds a small black handgun.\par
\textbf{MultiShotMaster.} \texttt{Story:} Subject 4: a man with grey hair; outfit: a grey suit, a light blue shirt, and a red striped tie. Subject 6: a woman with short brown hair; outfit: a zebra-patterned coat and dark pants. Subject 3: a large black mobile phone with an antenna. Subject 7: a small black handgun. Subject 5: an escalator. Setting: a lobby or hallway with an escalator. \texttt{Now:} Medium shot, low angle shot, static. Subject 6 is standing on the escalator. The woman raises a handgun and points it forward. By the end of the shot, Subject 6 holds Subject 7.\par
\textbf{HoloCine.} \texttt{[global caption]} [character2]: a man with grey hair; outfit: a grey suit, a light blue shirt, and a red striped tie. [character3]: a woman with short brown hair; outfit: a zebra-patterned coat and dark pants. Recurring object: a large black mobile phone with an antenna. Recurring object: a small black handgun. Setting: a lobby or hallway with an escalator. This scene contains 2 shots. \texttt{[per shot caption]} Medium shot, low angle shot, static. [character3] is standing on the escalator. The woman raises a handgun and points it forward. By the end of the shot, [character3] holds the small black handgun. \texttt{[shot cut]} \{shot 9\}\par
\textbf{StoryMem.} The setting is a lobby or hallway with an escalator. Medium shot, low angle shot, static. The woman with short brown hair is standing on the escalator. The woman with short brown hair is wearing a zebra-patterned coat and dark pants. The woman raises a handgun and points it forward. By the end of the shot, the woman with short brown hair holds the small black handgun.\par
\textbf{StoryDiffusion + Wan2.2-I2V.} Keyframe, SDXL: Medium shot, low angle shot. In a lobby or hallway with an escalator, a woman with short brown hair, is standing on the escalator, wearing a zebra-patterned coat and dark pants. Visible in the frame: a small black handgun. Video, I2V: the StoryMem text above.\par
\textbf{IC-LoRA + Wan2.2-I2V.} Grid panel, FLUX: Medium shot, low angle shot. In a lobby or hallway with an escalator, \textless Character 3\textgreater, is standing on the escalator, wearing a zebra-patterned coat and dark pants. Visible in the frame: a small black handgun. Video, I2V: the StoryMem text above.
\end{tcolorbox}
\end{PromptPlate}

\section{Video Samples}
\label{supp:videos}
\href{https://jinplu.github.io/CaT-Movie/}{The project-page videos} show eight test stories, one from each of eight films drawn at random with seed 42. Each video places \method, HoloCine, MultiShotMaster, and StoryMem side by side, the same videos Table~1 scores: the baselines render the reference storyboard and \method renders its planner's storyboard from the shot-list request. Each tile marks its own current shot, and the bar below shows the shot-list line of the current shot, timed by the reference storyboard (\cref{tab:supp_videos}).

\begin{table}[t]
  \centering
  \caption{\textbf{Stories in the videos on the project page.}}
  \label{tab:supp_videos}
  \TableSetup
  \begin{tabularx}{\columnwidth}{L C{1.1cm} C{1.1cm}}
    \toprule
    \TableHead
    Story & Shots & Seconds \\
    \midrule
    \texttt{cinedance-009077-b0004-shot01} & 12 & 36.3 \\
    \texttt{cinedance-010142-b0004-shot01} & 6 & 19.4 \\
    \texttt{cinedance-010691-b0000-shot01} & 7 & 39.2 \\
    \texttt{cinedance-010942-b0000-shot03} & 11 & 28.4 \\
    \texttt{cinedance-014083-b0000-shot01} & 12 & 21.8 \\
    \texttt{cinedance-014481-b0000-shot01} & 7 & 26.7 \\
    \texttt{cinedance-014600-b0000-shot01} & 7 & 20.2 \\
    \texttt{film-003995-b0001-shot06} & 12 & 27.8 \\
    \bottomrule
  \end{tabularx}
\end{table}

\section{Code}
\label{supp:code}
\href{https://github.com/JinPLu/CaT-Movie}{The code repository} contains the \texttt{mspc} package for data, annotation, training, inference, and evaluation, together with launch scripts, configurations, the adapted DiffSynth runtime, baseline pins and patches, and tests. \Cref{tab:supp_code} gives the entry point of each part. \texttt{uv sync -{}-locked} installs the locked environment, and configurations name data, model, and output locations under a workspace root set by \texttt{MSPC\_SHARED\_ROOT}.

\begin{table*}[t]
  \centering
  \caption{\textbf{Code entry points.} Paths are relative to the repository root.}
  \label{tab:supp_code}
  \TableSetup
  \begin{tabularx}{\textwidth}{l L}
    \toprule
    \TableHead
    Part & Entry point \\
    \midrule
    Shot screening and blocks (\cref{supp:data}) & \texttt{tools/data/preprocess\_sources.py}, \texttt{score\_shots.py}, \texttt{quality\_lines.py}, \texttt{build\_block\_members.py} \\
    Block annotation and spatial fields & \texttt{tools/data/annotate\_windows.py}, \texttt{validate\_annotations.py}, \texttt{spatial\_layer.py} \\
    Training windows and target latents & \texttt{tools/data/select\_training\_windows.py}, \texttt{build\_shot\_targets.py}, \texttt{run/precompute.sh} \\
    Renderer training (\cref{tab:supp_impl}) & \texttt{run/train.sh configs/train/renderer\_wan22\_final.py} \\
    Renderer ablations (Table~2) & \texttt{configs/train/renderer\_wan22\_final\_fixed\_role\_rows.py}, \texttt{renderer\_wan22\_final\_window\_shared.py}, and \texttt{renderer\_wan22\_final\_without\_references.py}: the final configuration without identity lines, routing, or references \\
    Planner (\cref{supp:planner}) & \texttt{tools/run/planner\_form\_plans.py plan -{}-prompt current -{}-sampling thinking-medium}, served by \texttt{run/vllm\_serve.py -{}-config configs/serving/vllm\_planner\_qwen38\_27b.yaml} \\
    Story rendering (Table~1) & \texttt{python -m mspc.benchmark.layout\_baseline write-planner-plans}, then \texttt{prepare}, both with \texttt{-{}-conditions configs/infer/planner\_renderer\_test92.json}, which reads \texttt{configs/infer/renderer\_wan22\_final\_step3000.py} \\
    Benchmark stories & \texttt{configs/benchmark/story\_requests\_v4.json}, \texttt{benchmark\_storyboards\_v4.json} \\
    Questions, judges, and scores (\cref{supp:benchmark}) & \texttt{src/mspc/evaluation/question\_writer.py}, \texttt{gemini\_checklist\_batch.py}, \texttt{consistency\_pair\_batch.py}, \texttt{tools/analysis/planned\_cut\_accuracy.py}; scores by \texttt{python -m mspc.evaluation.checklist\_scoring readout-v17}; text alignment in \texttt{renderer\_qualification\_public\_metrics.py} \\
    Paired intervals & \texttt{python -m mspc.evaluation.checklist\_scoring paired}, on \texttt{src/mspc/evaluation/paired\_bootstrap.py} \\
    Baselines (\cref{supp:baselines}) & \texttt{src/mspc/comparison\_methods/} \\
    Text-level planner analyses (\cref{supp:planner_text}) & \texttt{src/mspc/evaluation/planner\_state\_checks.py}, \texttt{shot\_text\_information.py} \\
    \bottomrule
  \end{tabularx}
\end{table*}

\FloatBarrier

\end{document}